\documentclass[final,5p,times,twoside,twocolumn]{elsarticle}
\usepackage{dblfloatfix}  

\usepackage[T1]{fontenc}
\usepackage[utf8]{inputenc}
\usepackage{amsmath,amssymb,amsfonts,amsthm}
\usepackage{booktabs}
\usepackage{graphicx}
\usepackage{xcolor}
\usepackage{enumitem}
\usepackage[protrusion=true,expansion=false]{microtype}
\usepackage{hyperref}
\usepackage{subcaption}
\usepackage{multirow}
\usepackage{array}
\usepackage{tabularx}
\usepackage{tcolorbox}
\usepackage{mdframed}
\usepackage{fancyvrb}
\usepackage{tikz}
\usetikzlibrary{arrows.meta,positioning,calc}
\usepackage{placeins}
\biboptions{numbers,sort&compress}

\definecolor{darkblue}{rgb}{0.0,0.1,0.4}
\definecolor{lightgray}{rgb}{0.95,0.95,0.95}
\definecolor{darkgreen}{rgb}{0.0,0.4,0.1}
\hypersetup{colorlinks=true,linkcolor=darkblue,citecolor=darkblue,urlcolor=darkblue}

\newcommand{\nh}{NH\textsubscript{4}}
\newcommand{\no}{NO\textsubscript{3}}
\newcommand{\nto}{N\textsubscript{2}O}
\newcommand{\doo}{O\textsubscript{2}}
\newcommand{\mcg}{\textsc{MCG}}
\newcommand{\drr}{\textsc{DRR}}
\newcommand{\Weff}{\mathbf{W}_\text{eff}}
\newcommand{\wwtp}{WWTP}
\newcommand{\bW}{\mathbf{W}}
\newcommand{\bp}{\mathbf{p}}

\newcommand{\pp}{\,pp}

\newmdenv[backgroundcolor=lightgray,linecolor=darkblue,linewidth=1pt,
          innerleftmargin=8pt,innerrightmargin=8pt,
          innertopmargin=6pt,innerbottommargin=6pt]{findingbox}

\definecolor{naivebg}{rgb}{0.98,0.95,0.95}
\definecolor{groundedbg}{rgb}{0.95,0.97,0.95}
\definecolor{ragbg}{rgb}{0.93,0.95,0.99}
\definecolor{naiveborder}{rgb}{0.7,0.3,0.3}
\definecolor{groundedborder}{rgb}{0.2,0.55,0.2}
\definecolor{ragborder}{rgb}{0.1,0.3,0.7}

\newmdenv[backgroundcolor=naivebg,linecolor=naiveborder,linewidth=1.2pt,
          innerleftmargin=7pt,innerrightmargin=7pt,
          innertopmargin=4pt,innerbottommargin=4pt,
          frametitle={\small\bfseries\color{naiveborder}Naive model},
          frametitleaboveskip=3pt]{naivebox}
\newmdenv[backgroundcolor=groundedbg,linecolor=groundedborder,linewidth=1.2pt,
          innerleftmargin=7pt,innerrightmargin=7pt,
          innertopmargin=4pt,innerbottommargin=4pt,
          frametitle={\small\bfseries\color{groundedborder}Grounded model},
          frametitleaboveskip=3pt]{groundedbox}
\newmdenv[backgroundcolor=ragbg,linecolor=ragborder,linewidth=1.2pt,
          innerleftmargin=7pt,innerrightmargin=7pt,
          innertopmargin=4pt,innerbottommargin=4pt,
          frametitle={\small\bfseries\color{ragborder}Grounded + RAG},
          frametitleaboveskip=3pt]{ragbox}

\begin{document}

\begin{frontmatter}

\title{Simulator-Grounded Large Language Models for Industrial
Causal Reasoning: Tool-Use, Structured Injection, and
Plant-Portable Retrieval for Wastewater Treatment Decision Support}

\author[aau]{Gary Simethy\corref{cor1}}
\ead{gasi@energy.aau.dk}
\author[aau]{Daniel Ortiz Arroyo}
\author[aau]{Petar Durdevic}

\cortext[cor1]{Corresponding author}
\affiliation[aau]{organization={Aalborg University, Department of Energy},
  addressline={Niels Bohrs Vej 8},
  city={Esbjerg},
  postcode={DK-6700},
  country={Denmark}}

\begin{abstract}
Wastewater operators need answers grounded in how their plant's
variables interact and how fast effects propagate, not in generic
pretraining text, when asking causal questions such as
\emph{``why is \nto{} rising?''} or \emph{``what happens if I cut
aeration by 20\%?''}.  We compare three concrete ways to ground a
frozen Qwen2.5-32B-Instruct model in an architecturally interpretable
wastewater simulator (CCSS-IX): a live simulator oracle (Method~1),
structured
parameter injection (Method~2), and a Decoupled Recall--Reasoning
(\drr{}) retriever (Method~3).  On a 198-question causal benchmark
the three reach \textbf{99.5\%}, \textbf{79\%}, and \textbf{75.8\%},
forming a deployment ladder above the strongest retrieval-augmented
baseline at $48\%$.  The \drr{} retriever has $110$M parameters and
trains per plant in $\sim\!17$ seconds; after cross-plant transfer to
a biologically distinct plant it still reaches \textbf{88\%}, while
Method~2's static table cannot transfer.  On a $60$-question
counterfactual benchmark only Method~3 handles queries about what
happens after an intervention: $+16.3\pp$ over Method~2, paired $95\%$
CI $[+7.1, +26.4]\pp$, with $100\%$ on the timescale and
operating-regime categories.  On the AI2 Reasoning Challenge (ARC)
with an OpenBookQA fact corpus, the same selective-retrieval mechanism
reaches \textbf{79\%} versus unconstrained Llama-3.1-8B $76\%$ and
full-injection $74\%$, a $+3\pp$ out-of-domain replication that argues
against a result specific to wastewater treatment.  We provide the
first single-simulator comparison of live tool-use, static parameter
injection, and learned numerical-parameter retrieval for industrial
causal question answering.
\end{abstract}


\end{frontmatter}

\section{Introduction}
\label{sec:intro}

Large language models (LLMs) have demonstrated strong capabilities
across scientific and technical domains~\citep{openai2023gpt4,
anthropic2024claude,google2023palm2}, yet their deployment in
safety-critical industrial settings requires more than fluent text
generation.  In a wastewater treatment plant (WWTP), a process engineer
asking ``Why is \nto{} (nitrous oxide, a potent greenhouse gas)
rising?'' or ``What happens if I reduce aeration by 20\%?'' needs
answers grounded in the
causal dynamics of the specific plant, not in statistical regularities
of internet text.  The gap between
general linguistic competence and domain-specific causal reasoning is
particularly acute in industrial process control, where variables interact
through nonlinear biochemical pathways, operating regimes shift over
timescales of minutes, and the cost of a wrong answer can be a regulatory
violation or a hazardous emission event.

\noindent\textbf{Why fine-tuning alone is the wrong tool.}\enspace The default move in language-model adaptation is supervised fine-tuning
(SFT) on a domain corpus.  But on industrial causal QA, SFT exhibits a
systematic \emph{recall--reasoning tradeoff}: the parametric memorisation
of plant facts (regime timescales, coupling weights, threshold rules)
improves, while the model's compositional reasoning over those same facts
degrades.  A growing empirical
literature~\citep{luo2023empirical,kotha2024understanding,biderman2024lora}
documents this tradeoff across domains and adapter ranks, and our own
198-question results (\S\ref{sec:experiments}) reproduce it: a model
fine-tuned on our Mechanistically-Grounded Corpus (\mcg{},
\S\ref{sec:mcg}) gains on recall-heavy categories but regresses on the
reasoning-heavy ones.  The present paper adopts the practical
conclusion of that pattern:
\emph{when an architecturally interpretable simulator of the target plant
exists, grounding the LLM in that simulator at inference time is strictly
preferable to fine-tuning the model parametrically on simulator outputs.}
Here we develop and compare three concrete grounding modes that realise
that strategy.

\noindent\textbf{Three grounding modes on a shared substrate.}\enspace All three modes share a single epistemic asset, the CCSS-IX
simulator~\citep{simethy2026epistemicengine}: an architecturally
interpretable open-loop \wwtp{} simulator that exposes per-timestep
regime identities $\bp_k(t)$, effective coupling matrices $\bW_k$,
eigenmode timescales, and a Causal Isolation Index for \nto{} spikes,
all derived from exact structural Jacobians through sparse coupling
matrices learned end-to-end.  The three modes differ in \emph{how} that
simulator's outputs reach the language model at inference time:
\begin{itemize}[leftmargin=1.6em,itemsep=2pt]
    \item \textbf{Method 1 -- Live Simulator Oracle.}{\raggedright
    The model retains its
    pretrained weights and is equipped with function-calling access to
    the live simulator: it issues tool queries
    (\texttt{run\_\allowbreak what\_\allowbreak if},
    \texttt{get\_\allowbreak timescale},
    \texttt{get\_\allowbreak coupling\_\allowbreak weight},
    \texttt{get\_\allowbreak coupling\_\allowbreak matrix},
    \texttt{get\_\allowbreak regime\_\allowbreak info}).}

  \item \textbf{Method 2 -- Structured Parameter Injection.}  The
    simulator is queried once \emph{offline} to extract a static parameter
    store; at inference time, a question-specific subset of parameters is
    prepended to the prompt of the frozen base model.  No runtime
    simulator, no fine-tuning, no retrieval training.
  \item \textbf{Method 3 -- The Decoupled Recall--Reasoning (\drr{})
    Architecture.}  A small plant-specific \emph{retriever} (a
    sentence-transformer bi-encoder, $\sim$110M parameters) is trained on
    (question, parameter) pairs from a Monte Carlo question generator,
    and selects the causally-relevant parameter subset that the frozen
    base model then conditions on.  Plant-conditioned retriever training fits in seconds per plant and the LLM is never updated.
\end{itemize}
The shared substrate matters: all three modes are validated on the same
198-question Causal Q\&A Benchmark, the same base model
(Qwen2.5-32B-Instruct), and the same simulator backend.  Differences in
accuracy and operational profile are therefore attributable to the
grounding mode itself, not to data, model, or simulator confounds.
To our knowledge, this is the first paper to compare live tool-use,
static parameter injection, and learned numerical-parameter retrieval
head-to-head \emph{for industrial causal QA} on a single benchmark
with a single base model.

\noindent\textbf{The deployment-ladder result.}\enspace On the 198-question Avedøre benchmark spanning all six causal-reasoning
categories:
\begin{itemize}[leftmargin=1.8em,itemsep=2pt]
  \item Live Simulator Oracle: \textbf{197/198 (99.5\%)}, with five of
    six categories at $100\%$; decisive where the runtime simulator
    dependency is acceptable.
    \item Structured Parameter Injection: \textbf{156/198 (79\%)},
    closing $\approx\!60\%$ of the static-corpus--oracle gap
    ($+30.8$ of $51.5\pp$ between the static-corpus baseline at
    $48\%$ and the oracle at $99.5\%$), with no fine-tuning and no
    runtime simulator.  Accuracy on the \emph{causal-edge} and
    \emph{regime} categories reaches $100\%$ and $97\%$ respectively
    (near oracle).

  \item \drr{} Architecture: \textbf{150/198 (75.8\%)} on the same 198Q
    using a hybrid (rule + learned) retriever, with the learned
    retriever trained on synthetic questions only (the $198$
    benchmark is held out from training); operationally close to
    Method 2 ($156/198$): both read question-specific parameters from
    the same store.  \drr{} is the more general option when the
    question-to-parameter mapping is not enumerable, and is the only
    mode that handles \emph{counterfactual} queries requiring
    post-intervention parameters (\S\ref{sec:cf-drr}).
\end{itemize}
Cross-plant transfer to a biologically distinct plant (Agtrup,
biological nutrient removal (BNR) configuration) reproduces the
spectrum: \drr{} reaches \textbf{35/40 (88\%)} after $\sim\!26$ seconds
of retriever training on $40$ gold-labelled questions plus $76$ Monte
Carlo synthetic questions; the LLM is never updated.  Cross-domain
transfer to the public AI2 Reasoning Challenge (ARC)
benchmark~\citep{clark2018arc} with an OpenBookQA knowledge corpus
reaches \textbf{79\%} overall ($316/400$ on a balanced $200$-Easy/$200$-Challenge
sample) with selective \drr{} retrieval, beating both unconstrained
base ($76\%$) and full-corpus injection ($74\%$); the modest but
consistent $+3\pp$ on this public out-of-domain benchmark is
consistent with a domain-general selective-retrieval effect, rather
than a WWTP-only artifact.

\noindent\textbf{Why all three modes, why not just the best one?}\enspace
The accuracy ranking $99.5\% > 79\% > 75.8\%$ is not the deployment
ranking; each mode has a distinct operational profile.  The oracle
requires both a running simulator and a tool-capable LLM at inference
time, so it belongs to a supervised control-room context.  Structured
injection requires neither, so it is the lightest deployment when the
question-to-parameter mapping is enumerable, but its static parameter
store cannot serve counterfactual queries.  \drr{} requires a small
per-plant retriever (trained from scratch in seconds while the LLM
stays untouched) and is the only non-oracle mode that handles
counterfactuals, via a simulator backend call routed by the retriever
rather than by the LLM (\S\ref{sec:cf-drr}).  A utility therefore
chooses by air-gap, simulator-availability, and LLM-capability
constraints, not by accuracy regret: within each constraint band the
corresponding method is on or near the empirical accuracy ceiling we
measured.  Table~\ref{tab:deployment} in \S\ref{sec:experiments}
summarises the requirements side by side with the accuracy results.

\medskip
\noindent\textbf{Contributions.}
\begin{enumerate}[leftmargin=1.8em,itemsep=4pt]
  \item \textbf{Three-mode comparison on a single mechanistic substrate.}
    Live tool-use, static parameter injection, and \drr{} (frozen base
    plus learned retriever) are evaluated on the same 198-question
    benchmark and CCSS-IX simulator, isolating where each mode wins
    (\S\ref{sec:experiments}).  Headline numbers: oracle \textbf{99.5\%},
    structured injection \textbf{79\%}, hybrid \drr{} \textbf{75.8\%}.
  \item \textbf{The \drr{} architecture and a plant-portable retriever.}
    A frozen Qwen-32B paired with a sentence-transformer bi-encoder
    retriever returns \emph{numerical parameters with physical context}
    (not text chunks).  Hybrid rule-plus-learned retrieval beats both
    components on Avedøre (Table~\ref{tab:retriever-ablation}); the
    architectural separation from static parameter injection appears
    on counterfactual queries (\S\ref{sec:cf-drr}).  Cross-plant
    transfer to Agtrup reaches $88\%$ in $\sim\!26$ seconds of
    retriever training (\S\ref{sec:drr}).
  \item \textbf{Cross-domain validation on a public benchmark.}
    The same selective-retrieval mechanism that drives \drr{} on
    \wwtp{} data, applied to ARC-Easy (recall) and ARC-Challenge
    (reasoning) with an OpenBookQA fact corpus, reaches
    $79\%$ vs base $76\%$ and full-injection $74\%$
    (\S\ref{sec:arc}); the gain is consistent with a domain-general
    selective-retrieval effect rather than a \wwtp{}-only artifact.
  \item \textbf{Mechanistically-Grounded Corpus generation that ports
    across plants (\mcg{}).}  A four-tier corpus pipeline (timestep,
    hourly, event, counterfactual) converts simulator outputs into LLM
    training records with no human annotation, and is validated as a
    \emph{baseline} for the three grounding modes
    (\S\ref{sec:mcg}--\ref{sec:crossplant}).
    \item \textbf{A 198-question causal benchmark and an evaluation audit.}
    We release the benchmark together with per-method evaluation outputs
    and bootstrap confidence intervals under both deterministic and
    semantic-judge scoring.  The paper describes a deterministic keyword
    scorer, an LLM-as-judge semantic layer, and a manual adjudication of
    all deterministic--semantic disagreements that traces $24/25$ to
    verdict-extraction failures rather than genuine disagreement
    (\S\ref{sec:bench}).

\end{enumerate}

\section{Related Work}
\label{sec:related}

\subsection{Language Models for Time Series and Industrial Data}

The question of how LLMs should interface with numerical time series
has attracted substantial recent attention.
LLMTIME~\citep{gruver2023llmtime} showed that pre-trained LLMs can
perform zero-shot forecasting when time series are serialised as
space-separated numerical strings, exploiting the tokenisation of
decimal numerals.  Time-LLM~\citep{jin2024timellm} introduced a
\emph{reprogramming} framework that patches time-series embeddings
into an LLM's representation space via an alignment layer.
PromptCast~\citep{xue2023promptcast} demonstrated that prompt-tuned
language models can compete with statistical forecasters on
univariate benchmarks.  One Fits All~\citep{zhou2023onefitsall} froze
a GPT-2 backbone and adapted only layer norm parameters to achieve
cross-domain generalisation.  Chronos~\citep{ansari2024chronos}
pre-trained a T5-family model on a curated time-series corpus,
achieving zero-shot forecasting without the numerical precision
issues of direct tokenisation.
More recent work has explored LLMs for industrial and scientific
time series specifically~\citep{liu2024itransformer,wang2024timexer,
sun2023test}.

A critical limitation shared by all of the above is that they model
\emph{statistical} associations in sensor streams.  None encodes
\emph{causal} process structure: which variable drives which, under
which regime, over what timescale, or what would happen under a
counterfactual control intervention.  Our three-method spectrum
addresses this gap by attaching the LLM (in three different ways) to
an architecturally interpretable simulator whose causal claims are
\emph{verified} rather than statistically approximated.

\subsection{Domain-Specific Fine-Tuning of LLMs}

The literature on domain-specific LLM adaptation has grown rapidly
across medicine~\citep{singhal2023medpalm,chen2023meditron},
law~\citep{cui2023chatlaw}, finance~\citep{xie2023pixiu}, and
code~\citep{roziere2023codellama}.  A consistent finding is that
domain fine-tuning improves task performance on tasks requiring
terminology, factual knowledge, or domain conventions
underrepresented in general pre-training~\citep{gururangan2020domains}.
Methodological cornerstones include
instruction-tuning~\citep{ouyang2022instructgpt},
chain-of-thought~\citep{wei2022chainofthought,kojima2022zeroshotthinkers,
wang2022selfconsistency}, and parameter-efficient adaptation through
low-rank adaptation (LoRA)~\citep{hu2022lora} and its quantised variant
QLoRA~\citep{dettmers2023qlora}, which we use throughout (rank
$r\!=\!16$, $\alpha\!=\!32$, on Qwen2.5-32B-Instruct~\citep{qwen2025qwen25}).

The critical question for domain LLMs is not \emph{whether} to
fine-tune but \emph{on what}.  For industrial process AI, no
publicly-available corpus is large or precise enough to absorb the
plant-calibrated parameters that mechanistic reasoning requires; the
structured causal knowledge lives in simulation models and expert
systems rather than in documents.  A growing literature also documents
a \emph{recall--reasoning tradeoff} under SFT on domain
QA~\citep{luo2023empirical,kotha2024understanding,biderman2024lora}:
domain SFT tends to gain on recall-style benchmarks while regressing
on the same model's compositional reasoning ability.
The practical conclusion (and the motivation for the present paper)
is that when an architecturally interpretable simulator is available,
grounding the LLM in that simulator at inference time (Methods 1--3)
is preferable to absorbing simulator outputs parametrically through
SFT.

\subsection{Retrieval-Augmented Generation and Frozen-Base Architectures}

Retrieval-augmented generation (RAG)~\citep{lewis2020rag} augments
LLM inference by retrieving relevant passages from an external
knowledge store and injecting them into the prompt.  RAG was initially
motivated
by the observation that parametric LLM memory is imprecise for factual
recall, while non-parametric retrieval can supply exact values on
demand~\citep{guu2020realm,izacard2021fid,shi2023replug}.  Large-scale
retrieval has been shown to close the factual-accuracy gap between
small and large models~\citep{lazaridou2022internet,
siriwardhana2023grounded,borgeaud2022retro}.

Our \drr{} architecture (\S\ref{sec:drr}) is a member of the
\emph{frozen-base + retriever} family but differs from text-RAG in
three architecturally important ways: (i) the corpus is numerical
parameters with physical context, not text chunks; (ii) retrieval is
supervised by causal relevance rather than free-text similarity; and
(iii) for counterfactual queries, the corpus is generated dynamically
by the simulator at query time, something a static text or
knowledge-graph corpus cannot do.  The closest published analogue
on the numerical-retrieval axis is
ARKNESS~\citep{cheng2025arkness}, which combines a knowledge graph
with retrieval to deliver numerically-exact answers for
computer-numerical-control manufacturing: ARKNESS demonstrates that
parameter retrieval beats text RAG, but its corpus is a static
knowledge graph, so it cannot serve counterfactual queries whose
answer depends on the post-intervention regime, which is exactly
the architectural moat our Cf-Bench experiment isolates
(\S\ref{sec:cf-drr}).  Frozen-base + plant-specific-adapter designs
in adjacent literatures include
Text-to-LoRA~\citep{charakorn2025t2l} and
Brainstacks~\citep{abuayyash2026brainstacks} (frozen
mixture-of-experts (MoE) LoRA stacks), but neither targets
industrial causal QA.
Our results illustrate a nuanced interaction between fine-tuning and
retrieval: the same retrieved parameter block, prepended to the
\emph{frozen} base model, beats fine-tuned (SFT) models that have to
absorb the same information parametrically.  RAG, in our setting,
\emph{amplifies} causal structure that the base model already
possesses; it does not create it.

\subsection{Synthetic and Simulator-Derived Corpus Generation}

The idea of generating training data from simulators or structured
knowledge sources predates the LLM era~\citep{wei2019eda,
sennrich2016back,chen2020kgpt}.  More recently,
\citet{wang2023selfinstruct} and \citet{alpaca2023} showed that LLMs
can generate instruction-following data;
\citet{gunasekar2023textbooks} (Phi-1) demonstrated that training on
synthetic ``textbook'' data can produce capable small models on coding
benchmarks.  In scientific ML, physics-informed
networks~\citep{raissi2019pinn}, neural
operators~\citep{li2021fourier,lu2021deeponet}, and physics foundation
models~\citep{herde2024poseidon,mccabe2023multiple} encode physical
laws as training constraints.  These approaches share with our
\mcg{} pipeline (\S\ref{sec:mcg}) the use of a structured
computational model to generate training data.  The distinction is
that \mcg{} bridges from \emph{numerical} simulator outputs to
\emph{natural language} training records, enabling an LLM to reason in
natural language about process dynamics that the simulator has already
verified.  In the present paper, the \mcg{} corpus serves as a
\emph{baseline} against which the three grounding modes are evaluated.

\subsection{Mechanistic Interpretability of Neural Models}

Mechanistic interpretability seeks to understand the computations
performed inside neural networks~\citep{olah2020zoom,
elhage2021mathematical,wang2022interpretability}.  Post-hoc attribution
methods such as SHAP~\citep{lundberg2017shap}, Integrated
Gradients~\citep{sundararajan2020axiomatic}, and
LIME~\citep{ribeiro2016lime} are known to be unstable under input
perturbations~\citep{ghassemi2021false} and cannot recover causal
structure not encoded in the model's
architecture~\citep{rudin2019stop}.  CCSS-IX's \emph{architectural}
interpretability sidesteps this concern: the coupling matrices and
response curves are not approximations derived after training; they
are the model's learned parameters, readable exactly.  Recent
LLM-interpretability work has identified induction
heads~\citep{olsson2022context}, knowledge-storage
components~\citep{meng2022rome}, and factual association
circuits~\citep{elhage2022superposition}; this lineage supports
the broader observation that fine-tuning a base model perturbs
specific compositional circuits, which in turn motivates the
frozen-base preference of Methods 2--3.

\subsection{LLMs for Scientific and Engineering Reasoning}

Several recent efforts have targeted scientific reasoning:
Galactica~\citep{taylor2022galactica} on scientific literature,
Med-PaLM~\citep{singhal2023medpalm} on USMLE-style questions,
Minerva~\citep{lewkowycz2022minerva} on quantitative reasoning,
MechGPT~\citep{buehler2024mechgpt} on materials science.  None
directly studies industrial process control, where the relevant causal
knowledge is distributed across simulator state trajectories, control
schedules, and biochemical process models rather than in scientific
literature.

\subsection{Wastewater Treatment Process AI}

Activated Sludge Models (ASM1, ASM2,
ASM3)~\citep{henze1987asm1,henze2000asm} provide first-principles
descriptions of biological nitrogen and phosphorus removal and have been used for process design, optimisation, and fault
detection~\citep{copp2002benchmark,alex2008benchmark,
jeppsson2006benchmark}.  Machine-learning approaches to \wwtp{}
modelling include effluent prediction~\citep{bagheri2015nnwwtp,
qiao2021lstm}, energy
optimisation~\citep{corominas2018ai}, hybrid mechanistic-data
models~\citep{boger2006hybrid}, and physics-informed
losses~\citep{newhart2019hybrid}.
LLM work for water/wastewater is beginning to emerge.
WaterER~\citep{xu2025waterer} is a 1{,}043-task
evaluation suite; WaterGPT~\citep{xu2025watergpt} argues for domain
adaptation and tool augmentation; text-retrieval RAG over
operational records and standard operating procedures (SOPs) has
been explored as the natural baseline against which our \drr{}
parameter-retrieval result is positioned, but lacks access to
plant-calibrated numerical couplings.  Multi-agent decision-support
frameworks~\citep{rothfarb2025multiagent} and LLMs augmented with
data-driven small models~\citep{xu2025smallmodels} have also
appeared.  Recent \emph{Making Waves} surveys outline a research
agenda for \wwtp{} LLM-agent systems~\citep{xu2025makingwaves_agents}
and multi-agent water-engineering decision
support~\citep{hosseini2025makingwaves_multi}.  Across this emerging line,
the adaptation is textual rather than mechanistically grounded in a
simulator.  This paper contributes the first single-simulator
comparison of grounding modes for industrial causal QA within that
agenda.

\noindent\textbf{LLM + simulator for water distribution.}\enspace A closely-related water-engineering line couples LLMs to the
\textsc{EPANET} hydraulic simulator for water distribution systems.
\citet{goldshtein2025llmepanet} build an agent that issues
\textsc{EPANET} tool calls with retrieval augmentation and reports
$56$--$81\%$ accuracy on natural-language queries about
distribution-network state; \citet{wang2026epanetagentic} extend
this to a multi-agent orchestrator with $100\%$ tool-invocation
accuracy on L-Town, C-Town, and Net3 networks;
\citet{wen2026wateradmin} pair an LLM operating-mode selector
with a downstream optimiser for community-scale water distribution.
These works share our architectural pattern (frozen LLM + simulator
tool calls + retrieval) but target water-distribution control,
where the simulator is mechanistic (hydraulic equations) rather
than learned from sensor data, and the question class is operational
recommendation rather than \emph{causal} QA about post-intervention
plant dynamics.  The deployment-spectrum framing here is complementary: our three-mode
comparison on a single substrate could be applied to the
\textsc{EPANET}-agentic line by adding static parameter injection and
learned-retriever variants alongside the existing tool-calling agents.

Three concurrent works are architecturally adjacent.
\citet{vyas2025autonomous} build an autonomous control framework in
which an LLM agent plans, queries a process simulator, and validates
its actions against a physical setpoint, a control-loop sibling of
our oracle paradigm (Method 1) but evaluated on plant-wide control
rather than natural-language causal QA.
\citet{kampourakis2026dtics} couple a digital twin of a water-treatment
process with a JSON-schema-constrained LLM for cyber-physical anomaly
detection; the architectural trio (twin + structured-output LLM +
schema) is the closest published analogue of our oracle pipeline, but
the question class is security-incident detection.
Tool-use frameworks such as ChemCrow~\citep{bran2024chemcrow} and
Coscientist~\citep{boiko2023coscientist} for chemistry, together with
the general Toolformer/ReAct
lineage~\citep{schick2023toolformer,yao2023react}, demonstrate that
function-calling with frontier LLMs can drive scientific workflows,
but none compares against fine-tuned domain models on a single shared
benchmark.

Our application contribution is therefore narrower but more
operational: rather than proposing a generic wastewater assistant, we
study how an interpretable \wwtp{} simulator can be exposed as
\emph{three} deployment modes for plant-specific causal QA, then
validate those modes on Avedøre and Agtrup, and on the public ARC
benchmark for domain generality.

\subsection{Reinforcement Learning and Self-Improvement for LLMs}

Recent LLM training has moved beyond SFT toward reinforcement-based
self-improvement: RLHF~\citep{ouyang2022instructgpt}, Constitutional
AI~\citep{bai2022constitutional}, DeepSeek-R1~\citep{deepseek2025r1}
with GRPO~\citep{shao2024deepseekmath}, where a verifiable oracle
(symbolic correctness, ground-truth labels) replaces the human reward
signal.  For our setting, the CCSS-IX simulator is a
physically-grounded oracle: a counterfactual trajectory predicted by
the LLM can be verified by running the simulator under the specified
control perturbation.  The \mcg{} corpus and the 198-question
benchmark in this work provide the foundation for any future
RL-based adaptation of the spectrum.

\section{Background and Problem Setting}
\label{sec:background}

\subsection{CCSS-IX as an Epistemic Simulator}
\label{sec:ccss_ix}

All three grounding modes studied in this paper share an upstream
simulator.  We use CCSS-IX (Continuous-time Compositional State-Space,
interpretable variant)~\citep{simethy2026epistemicengine}, a
continuous-time regime-switching state-space model for open-loop
\wwtp{} simulation, because it provides \emph{exact} mechanistic
outputs rather than approximations.

CCSS-IX extends the black-box CCSS-RS (regime-switching scaffold)
ancestor~\citep{simethy2026ccssrs} by replacing opaque MLP fast
experts with structured experts
parameterised by a sparse structural coupling matrix
$A_k^{(0)}\!\in\!\mathbb{R}^{D \times D}$ and per-variable
one-dimensional response curves $g_i^{(k)}$, both trained end-to-end.
The key architectural property is that the structural Jacobian
$\partial f_k / \partial x$ equals $A_k^{(0)}$ exactly (modulo a
nonlinear correction from $g_i^{(k)}$), enabling \emph{exact} causal
coupling extraction without post-hoc approximation.

CCSS-IX provides four structural outputs that constitute the shared
epistemic foundation for our three modes:
\begin{enumerate}[leftmargin=1.8em,itemsep=2pt]
  \item $\mathbf{p}_k(t)\!\in\![0,1]^K$: regime posterior
    probabilities, giving instantaneous regime identity (three
    regimes: \emph{aerobic-fast}, oxygen present and fast
    nitrification; \emph{standard aerobic-anoxic}, mixed conditions;
    and \emph{slow-anoxic}, oxygen-depleted and denitrification
    dominant).
  \item $W_k\!\in\!\mathbb{R}_{\geq 0}^{D \times D}$: effective
    coupling weight matrices per regime, derived from the structural
    Jacobians, encoding the strength and direction of inter-variable directed influence.
  \item $\tau_{kd}$: per-variable eigenmode timescales per regime,
    characterising decay rates of perturbations to each state variable.
  \item $\text{CII}(t)$: the Causal Isolation Index, a scalar that
    measures the degree to which current \nto{} dynamics are driven by
    external state-variable coupling vs.\ autonomous internal
    dynamics.
\end{enumerate}

Validation of these outputs against biochemical ground truth is
documented in~\citet{simethy2026epistemicengine}: $6/8$ ASM1
biochemical causal edges are recovered from $W_k$ at the
$5\%$-of-row-maximum threshold (the remaining two recover at slightly
higher thresholds), timescale ratios between regimes align with known
aerobic/anoxic kinetics, and CII provides $\sim\!94$-minute mean lead
time on coupled \nto{} spikes with a $90.3\%$ ($112/124$)
coupled-event labelling rate.

\subsection{Problem Statement}
\label{sec:problem}

Given the CCSS-IX simulator with parameters $\theta^*$ trained on an
industrial \wwtp{} dataset $\mathcal{D}$, and a target LLM
$\mathcal{M}$ with pre-trained weights $\phi_0$, the three grounding
modes introduced in \S\ref{sec:intro} are formally as follows.

\paragraph{Method 1 -- Live Oracle}
Keep $\phi\!=\!\phi_0$; expose the five CCSS-IX function-calling
tools defined in \S\ref{sec:oracle-tools} to $\mathcal{M}$.

\paragraph{Method 2 -- Structured Parameter Injection}
Keep $\phi\!=\!\phi_0$; extract the static parameter store
$\mathcal{P}(\theta^*) = \{W_k, \tau_{kd}, \text{thresholds}, \dots\}$
offline, and prepend a hand-crafted question-specific subset
$\mathcal{P}_Q\!\subseteq\!\mathcal{P}$ to the prompt.

\paragraph{Method 3 -- Decoupled Recall--Reasoning (\drr{})}
Keep $\phi\!=\!\phi_0$; train a sentence-transformer bi-encoder
retriever $R_\theta$ ($\sim\!110$M parameters) on
$(Q, \mathcal{P}_Q)$ pairs from a Monte Carlo generator over
$\mathcal{P}$, and set $\mathcal{P}_Q = R_\theta(Q)$.

\paragraph{Mode 0 -- Corpus-grounded fine-tuning (baseline)}
Fine-tune $\mathcal{M}$ on a corpus
$\mathcal{C}(\theta^*, \mathcal{D})$ derived from the simulator
outputs:
\[
\hat{\phi} = \arg\max_\phi \; \mathbb{E}_{(Q,A^*) \sim \mathcal{B}} \big[
  \log P_\phi(A^* \mid Q) \big],
\]
where $\mathcal{B}$ is the 198-question Causal Q\&A Benchmark and
$\mathcal{C}$ is the \mcg{} corpus (\S\ref{sec:mcg}).  The
recall--reasoning tradeoff~\citep{luo2023empirical,
kotha2024understanding,biderman2024lora} that motivates frozen-base
Modes 2--3 over Mode 0 is well documented in the literature and is
not the contribution of this paper.

\section{The \mcg{} Pipeline (Baseline Mode 0)}
\label{sec:mcg}

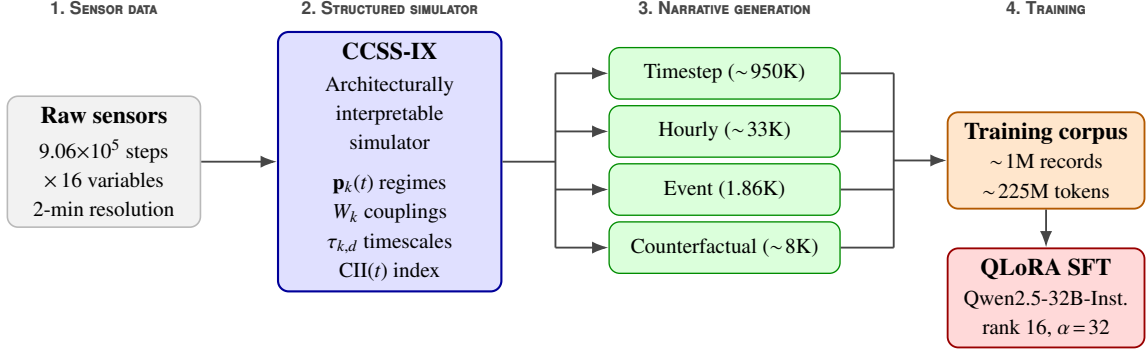
\begin{figure*}[t]
\centering
\begin{tikzpicture}[
  font=\small,
  every node/.style = {align=center},
  stage/.style      = {font=\scriptsize\sffamily\bfseries, gray!55!black},
  rawbox/.style     = {draw=gray!55, fill=gray!10, rounded corners=4pt,
                       minimum width=2.55cm, minimum height=1.7cm,
                       line width=0.6pt, inner sep=4pt},
  ixbox/.style      = {draw=blue!60!black, fill=blue!12, rounded corners=4pt,
                       minimum width=2.95cm, minimum height=3.15cm,
                       line width=0.8pt, inner sep=4pt},
  narrbox/.style    = {draw=green!55!black, fill=green!14, rounded corners=4pt,
                       minimum width=3.05cm, minimum height=0.68cm,
                       font=\footnotesize, line width=0.5pt, inner sep=3pt},
  corpusbox/.style  = {draw=orange!70!black, fill=orange!20, rounded corners=4pt,
                       minimum width=2.6cm, minimum height=1.2cm,
                       line width=0.7pt, inner sep=4pt},
  sftbox/.style     = {draw=red!60!black, fill=red!15, rounded corners=4pt,
                       minimum width=2.6cm, minimum height=1.2cm,
                       line width=0.7pt, inner sep=4pt},
  arrow/.style      = {-Latex, line width=0.7pt, gray!55!black},
  bus/.style        = {line width=0.7pt, gray!55!black},
]
  \node[rawbox] (raw) at (0, 0) {%
    \textbf{Raw sensors}\\[2pt]
    \footnotesize $9.06{\times}10^{5}$ steps\\
    \footnotesize $\times\,16$ variables\\
    \footnotesize 2-min resolution};

  \node[ixbox, right=1.0cm of raw] (ix) {%
    \textbf{CCSS-IX}\\[1pt]
    \footnotesize Architecturally\\
    \footnotesize interpretable\\
    \footnotesize simulator\\[4pt]
    \footnotesize $\bp_k(t)$~regimes\\
    \footnotesize $W_k$~couplings\\
    \footnotesize $\tau_{k,d}$~timescales\\
    \footnotesize CII$(t)$~index};

  \node[narrbox, anchor=west]
        at ([xshift=1.4cm, yshift=0.40cm]ix.east) (n2)
        {Hourly $(\sim\!33$K$)$};
  \node[narrbox, above=2pt of n2] (n1) {Timestep $(\sim\!950$K$)$};
  \node[narrbox, below=2pt of n2] (n3) {Event $(1.86$K$)$};
  \node[narrbox, below=2pt of n3] (n4) {Counterfactual $(\sim\!8$K$)$};

  \coordinate (narrCent) at ($(n2.east)!0.5!(n3.east)$);
  \node[corpusbox, right=1.4cm of narrCent] (corpus) {%
    \textbf{Training corpus}\\
    \footnotesize $\sim\!1$M records\\
    \footnotesize $\sim\!225$M tokens};
  \node[sftbox, below=0.5cm of corpus] (sft) {%
    \textbf{QLoRA SFT}\\
    \footnotesize Qwen2.5-32B-Inst.\\
    \footnotesize rank $16$, $\alpha\!=\!32$};

  \coordinate (labelBar) at ([yshift=0.30cm]ix.north);
  \node[stage] at (raw    |- labelBar) {\textsc{1.\ Sensor data}};
  \node[stage] at (ix     |- labelBar) {\textsc{2.\ Structured simulator}};
  \node[stage] at (n1     |- labelBar) {\textsc{3.\ Narrative generation}};
  \node[stage] at (corpus |- labelBar) {\textsc{4.\ Training}};

  \draw[arrow] (raw.east) -- (ix.west);

  \coordinate (busL) at ([xshift=-0.7cm]n1.west);
  \draw[bus]   (ix.east) -- (busL |- ix.east);
  \draw[bus]   (busL |- n1) -- (busL |- n4);
  \draw[arrow] (busL |- n1) -- (n1.west);
  \draw[arrow] (busL |- n2) -- (n2.west);
  \draw[arrow] (busL |- n3) -- (n3.west);
  \draw[arrow] (busL |- n4) -- (n4.west);

  \coordinate (busR) at ([xshift=-0.7cm]corpus.west);
  \draw[bus]   (n1.east) -- (busR |- n1);
  \draw[bus]   (n2.east) -- (busR |- n2);
  \draw[bus]   (n3.east) -- (busR |- n3);
  \draw[bus]   (n4.east) -- (busR |- n4);
  \draw[bus]   (busR |- n1) -- (busR |- n4);
  \draw[arrow] (busR) -- (corpus.west);

  \draw[arrow] (corpus.south) -- (sft.north);
\end{tikzpicture}
\caption{The \mcg{} pipeline converts architecturally-verified
  simulator outputs ($p_k$, $W_k$, $\tau$, CII) into four complementary
  narrative types and assembles a $\sim\!225$M-token training corpus,
  used for QLoRA fine-tuning of Qwen2.5-32B (Mode 0 baseline).  The
  same simulator outputs are also exposed as the static parameter store
  for Method 2 (\S\ref{sec:struct}) and the retriever corpus for
  Method 3 (\S\ref{sec:drr}), and as live tools for Method 1
  (\S\ref{sec:oracle}).}
\label{fig:pipeline}
\end{figure*}

The \mcg{} (Mechanistically-Grounded Corpus) pipeline
(Figure~\ref{fig:pipeline}) is the \emph{corpus-grounded
fine-tuning} baseline against which the three inference-time
grounding modes are evaluated.  We summarise it here; the
construction details that follow apply equally to the \emph{static
parameter store} that Methods 2--3 later draw from.

\subsection{Signal Extraction}
\label{sec:extraction}

For each contiguous 1{,}000-step window $w$ in the Avedøre \wwtp{}
dataset (906{,}000 timesteps, 16 variables, 2-minute resolution), we
run CCSS-IX in open-loop mode to extract
\[
\text{signals}(w) = \big\{\hat{y}_{w,t},\, \bp(w,t),\, W_k,\, \tau_{kd},\,
   \text{CII}_z(w,t),\, \text{top-}K\text{ edges at }t \big\}_{t=0}^{H-1}
\]
where $\hat{y}_{w,t}\!\in\!\mathbb{R}^D$ are predicted state values in
physical units, $\bp(w,t)\!\in\!\Delta^{K-1}$ are regime posteriors,
and top-$K$ edges are the $K$ largest entries of
$W_{\text{dominant}(t)}$ above a minimum strength threshold.  For
Avedøre, $D\!=\!5$ (\nh{}, \no{}, \nto{}, \doo{}, suspended
solids (SS)), $K\!=\!3$
regimes, and $1{,}000$ windows of $H\!=\!1{,}000$ steps yield
$10^{6}$ timestep signal tuples.  \nto{} spike detection identifies
$1{,}857$ events of which $420$ ($22.6\%$) are coupled
(CII$_z\!>\!1.0$ in the 120-minute lookback).

\subsection{Four Narrative Generators}
\label{sec:narratives}

\textbf{Timestep narratives} ($\sim\!950{,}000$ records): for each
sampled timestep, a Q\&A pair describing the current operating regime,
dominant coupling edges, variable states, and CII value.
\textbf{Hourly narratives} ($\sim\!33{,}000$ records): 60-minute
windows summarised with regime-occupancy statistics, dominant coupling
changes, and trend directions.
\textbf{Event narratives} ($1{,}857$ records): for each detected
\nto{} spike, a causal-attribution narrative classifying the spike as
coupled (CII-driven, $420$ events) or isolated, with the precursor
window and expected lead time; the $\sim\!94$-minute mean
early-warning lead is encoded in every coupled-event record.
\textbf{Counterfactual Q\&A} ($\sim\!8{,}000$ records): for each
sampled intervention timestep, CCSS-IX is re-run with a modified
control input ($\pm 20\%$ on \doo{}.SETPOINT, FLOW, TEMP) and the
predicted trajectory delta is expressed as an interventional Q\&A pair;
counterfactual pairs are the most informationally dense type and
receive the highest sampling weight during fine-tuning.

\subsection{Fact-Card Injection}
\label{sec:factcards}

A pilot fine-tune revealed a systematic failure mode: the model
could describe qualitative causal structure accurately but
hallucinated quantitative constants.  We address this with
\emph{fact cards}: a JSONL of $64$ atomic question-answer pairs, each
testing a single numeric constant (timescales, $W_{\text{eff}}$ values,
regime usage fractions, CII threshold/lead time, counterfactual
horizon, ASM1 edge count).  Fact cards are generated programmatically
from the ground-truth $W_k$ and $\tau_{kd}$ matrices and oversampled
$20\times$ during fine-tuning, yielding $1{,}280$ injected records
alongside $50{,}000$ corpus records.

\subsection{Fine-Tuning}
\label{sec:finetuning}

We fine-tune Qwen2.5-32B-Instruct~\citep{qwen2025qwen25} using QLoRA
(rank $16$, $\alpha\!=\!32$, $\eta\!=\!2\cdot 10^{-4}$, batch $4$ with
grad-accum $\times 8$, $2$ epochs, scaled dot-product attention) on
$50{,}000$
sampled corpus records plus the oversampled fact-card set.  Final
training loss $0.2325$ and token accuracy $92.2\%$ over $3{,}206$ steps
($\sim\!13.6$ hours on a single NVIDIA RTX PRO 6000 Blackwell (96\,GB)).

\subsection{Naive Corpus Baseline}
\label{sec:naive}

To isolate the contribution of mechanistic grounding, we construct a
\emph{naive corpus} using the same four narrative types but replacing
all mechanistic content with raw sensor statistics (values, means,
standard deviations, trend directions).  The naive corpus does
\emph{not} use regime identities, coupling matrices, timescales, CII,
or counterfactual trajectories.  It is a high-quality sensor-text
template corpus, analogous to prior \wwtp{} NLP
work~\citep{bagheri2015nnwwtp}.  Fine-tuning the same base model on
the same 50{,}000-record sample yields the \emph{naive model} for
ablation (final loss $\sim\!0.326$, token accuracy $\sim\!87.2\%$). Causal-QA accuracy of
the resulting naive model is reported alongside the three Grounded
variants in Table~\ref{tab:headline} and analysed in the case studies
of \S\ref{sec:cases}.

\subsection{Retrieval-Augmented Inference (Mode 0+RAG)}
\label{sec:rag}

For the grounded+RAG condition we prepend a static structured
knowledge header (per-regime timescales, top-$K$ $W_{\text{eff}}$
edges per regime, CII threshold/lead-time constants, regime usage
fractions, ASM1 edge count) to the system prompt at inference time.
This is a non-parametric injection of the same information that is
parametrically attempted through fact-card fine-tuning; the header is
$\sim\!400$ tokens and identical across questions.  A
\emph{regime-aware RAG} variant detects the implied operating regime
from the question text (lexical and physiochemical cues) and prepends a
focused excerpt before the full header.  Mode 0 + RAG is the
strongest static-injection variant of the corpus-grounded paradigm and
is the most challenging baseline for Methods 2--3.

\section{Method 1: The Simulator Oracle}
\label{sec:oracle}

Method 1 is the most simulator-coupled of the three modes: the base
LLM (Qwen2.5-32B-Instruct, no fine-tuning) is equipped with live
CCSS-IX tool calls via function calling.  The model acts as an agent,
issuing tool queries to the running simulator and synthesising causal
answers from the numerical responses.  No training signal from the
benchmark is used; the model's causal competence comes entirely from
its ability to query and interpret the CCSS-IX oracle at runtime.

\subsection{Tool Interface}
\label{sec:oracle-tools}

The model can ask the simulator five things at inference time:
what would change under a control perturbation, how fast a state
variable responds, how strongly two variables couple, what the full
coupling matrix looks like for a regime, and what regime metadata and
CII early-warning constants apply.  Each is exposed as a
function-calling tool:
\begin{description}[leftmargin=1.8em,itemsep=3pt]
  \item[\texttt{run\_what\_if(variable, delta\_pct, regime)}]
    Simulate a counterfactual perturbation: run CCSS-IX under the
    specified control variable change and return the mean $\Delta y$
    per state variable over the simulator horizon.
  \item[\texttt{get\_timescale(variable, regime)}]
    Return $\tau_{\text{variable}}$ in minutes for the given operating
    regime, derived from the CCSS-IX eigenmode decomposition.
  \item[\texttt{get\_coupling\_weight(source, target, regime)}]
    Return $W_{\text{eff}}(\text{source}{\to}\text{target})$ in the
    specified regime, extracted from the learned structural coupling
    matrix.
  \item[\texttt{get\_coupling\_matrix(regime, top\_k)}]
    Return the full $W_{\text{eff}}$ matrix for a regime together
    with the top-$k$ strongest edges, used for multi-hop chain and
    ranking questions.
  \item[\texttt{get\_regime\_info()}]
    Return regime metadata (names, occupancy fractions, descriptions)
    and CII early-warning constants (alert threshold, lead-time
    percentiles, coupling rate, and total spike count).
\end{description}

\subsection{Agentic Loop}
\label{sec:oracle-loop}

The model issues $1$--$5$ tool calls per question, then synthesises a
causal answer from the accumulated numerical outputs.  The loop is:
(1) the model receives the question and available tool schemas;
(2) it issues a tool call (or chain of calls) to gather relevant
CCSS-IX outputs; (3) it synthesises a natural-language answer grounded
in the tool responses.

After step (3), the draft answer is checked for \emph{forbidden
concepts}: per-question wrong-answer phrasings curated in the
benchmark alongside the gold key (for example, naming the wrong
causal direction, the wrong spike classification, or a rejected
alternative the question explicitly contrasts against).  The check
uses the same word-level matching with negation awareness as the
deterministic scorer (\S\ref{sec:bench}), so a contrastive phrasing
such as \emph{``classified as coupled, not isolated''} does not fire
on \emph{isolated}.  When a forbidden hit is found, one
\emph{self-correction turn} is triggered: the model is told that the
matched phrasing should not be named and asked for a revision that
states only the correct alternative (without re-naming the rejected
one).  The mechanism uses benchmark-side information (the curated
forbidden list) as a runtime correction signal, an agent-loop
self-refinement step whose contribution we report separately: in the
headline ablation (\S\ref{sec:experiments}) this turn adds $+4\pp$
over the tools-only $189/198$ ($95.5\%$), flipping the eight
self-corrected questions from $0/8$ to $8/8$ to reach the headline
$197/198$ ($99.5\%$).

\subsection{System Prompt and Domain Rules}
\label{sec:oracle-prompt}

The oracle system prompt encodes a handcrafted rule set covering three
categories:
\begin{itemize}[leftmargin=1.8em,itemsep=2pt]
  \item \textbf{Biochemistry direction overrides}: explicit rules for
    which direction each variable should move under each control
    perturbation in each regime, preventing the generic LLM prior from
    overriding simulator outputs (e.g., \emph{``under reduced \doo{}
    in regime $k\!=\!2$, \nh{} \emph{increases} because nitrification
    is already \doo{}-limited and ammonia-oxidising bacteria (AOB)
    activity becomes diffusion-limited inside flocs''}).
    \item \textbf{Vocabulary and surface-form mandates.}\enspace Beyond a
    generic $9$-rule tool-use methodology, the oracle prompt includes
    additional rules that prescribe required terminology and exact
    phrasings for specific question patterns (e.g.\ classification
    labels, verbatim wording for sign-of-change phrases, and the
    explicit \emph{``coupled / isolated''} dichotomy tied to the
    CII $z$-score threshold; for the z-score above $2.0$ case the
    prompt requires the word \emph{``coupled''}, and below threshold
    forbids it).  These rules target the deterministic scorer's
    required- and forbidden-keyword sets so that a correct internal
    classification is also rendered in the surface form the scorer
    accepts. The full $57$-rule prompt is included in the data repository
  referenced in \S\ref{sec:data}, and the contribution
  of the extended rule set ($+22.3\pp$ over the $9$-rule baseline)
  is reported in the headline ablation (\S\ref{sec:experiments}).

  \item \textbf{Causal chain templates}: structured reasoning templates
    that fix the chain skeleton for multi-hop answers (e.g., the
    four-step \doo{}.SETPOINT-to-\nto{} chain must spell out
    \emph{``$O_2 \uparrow \to$ AOB active $\to$ \nh{} converted to
    \no{} (\nh{} decreases) $\to$ \nto{} suppressed''}, explicitly
    naming \nh{} as the intermediate).
\end{itemize}
The system prompt used in our experiments contains $57$ numbered
rules, grown from a generic $9$-rule pilot through inspection of
failure cases.  The prompt does not embed any numerical constants; all
quantitative information comes from the live tool calls.  This
separation ensures that the oracle's accuracy derives from CCSS-IX,
not from prompt engineering of factual values, though prompt
engineering does shape how the model interprets those facts: the
$9$-rule $\to$ $57$-rule upgrade alone contributes $+22.3\pp$
(~\ref{app:oracle-ablation}, $145/198 \to 189/198$, both
without self-correction), making the rule set the largest single
contributor to Method 1's headline score.

\subsection{Evaluation Protocol}
\label{sec:oracle-eval}

Method 1 is evaluated on the full 198-question benchmark
($6$ categories $\times$ $33$ questions), scored with the same
deterministic keyword-based scorer used for the corpus baselines:
each response must match every required keyword set for the
question and trigger none of its forbidden phrasings, with
negation-aware matching so a contrastive sentence such as
\emph{``classified as coupled, not isolated''} does not falsely
fire on \emph{``isolated''} (\S\ref{sec:bench}).  Partial-match
cases (some required keywords missing, no forbidden hit) are
re-scored by a DeepSeek-R1-Distill-Qwen-14B semantic auditor;
the manual adjudication of all deterministic--semantic
disagreements (~\ref{app:adjudication}) traces $24/25$ to
verdict-extraction failures rather than genuine disagreement.
Targeted ablations isolating the prompt's contribution and the
self-correction loop's contribution appear in the headline-ablation
discussion of \S\ref{sec:headline} and the breakdown table in ~\ref{app:oracle-ablation}.

\section{Method 2: Structured Parameter Injection}
\label{sec:struct}

\noindent\textbf{Motivation.}\enspace Method 1 achieves near-perfect accuracy ($197/198$, $99.5\%$) but
requires a live CCSS-IX process at inference time.  The corpus
baseline plateaus at $95/198$ ($48\%$) because its static knowledge
header supplies aggregate context rather than question-specific
parameters.  We ask: can oracle-like accuracy be recovered
\emph{without} the runtime simulator by replacing the static header
with a question-specific block assembled from the same parameter
store?  We term this approach \emph{structured parameter injection}.

\noindent\textbf{Method.}\enspace At inference time we parse each question to detect its type
(causal\_edge / regime / anomaly / early\_warning / counterfactual /
multi\_hop) and extract the relevant plant entities (source variable,
target variable, operative regime).  A lightweight lexical lookup
retrieves the exact CCSS-IX values for those entities and serialises
them into a compact block.  For a causal-edge question
\emph{``What is the coupling weight from \no{} to \nto{} in the
aerobic-fast regime?''}, the block is:

\begin{Verbatim}[fontsize=\small]
[Causal parameters: NO3 -> N2O]
W_eff(NO3->N2O, k=0 aerobic-fast) = 0.312
W_eff(NO3->N2O, k=1 standard)     = 0.274
W_eff(NO3->N2O, k=2 slow-anoxic)  = 0.241
Strongest driver of N2O: NO3 (0.312 in k=0)
\end{Verbatim}

\noindent For regime questions the block contains the complete
timescale table; for anomaly and early-warning, CII thresholds and
lead-time statistics; for counterfactual and multi-hop, the full
$W_k$ matrix and timescales for the relevant regime.  This $200$--$400$
token structured block replaces the $1{,}200$-token prose knowledge
header, reducing context size while increasing precision.  Critically,
\emph{no simulator process is running}: the block is assembled in
microseconds from pre-computed static parameters, matching corpus-model
latency while providing oracle-quality facts.

We evaluate two structured conditions.
\textbf{Base+Struct}: untuned Qwen2.5-32B-Instruct with structured
injection (no fine-tuning).
\textbf{Grounded+Struct}: Avedøre SFT checkpoint with structured
injection.  Method 2's headline number reported in the abstract
($156/198$, $79\%$) is Base+Struct.

\noindent\textbf{Where Method 2 wins, and where it stops.}\enspace
Per-category, Method 2 reaches oracle accuracy on factual-lookup
questions (causal edge $100\%$, regime $97\%$; full breakdown in
\S\ref{sec:experiments}) and trails the oracle on the compositionally
harder categories (multi-hop, counterfactual).  The architectural
ceiling on counterfactuals is structural: the static store contains
only \emph{pre-intervention} parameters, so for a question that asks
what happens \emph{after} a $\pm 20\%$ control perturbation, the
required post-intervention $W_k$ does not exist in $\mathcal{P}$ and
no lookup can supply it.  Method 3 (\S\ref{sec:drr}) lifts this
ceiling by re-running CCSS-IX on demand and routing the
post-intervention parameters through a trained retriever.

\noindent\textbf{Why Method 2 closes the gap.}\enspace The \mcg{}-fine-tuned model (Mode 0, Grounded) absorbs the same
parameters as the structured block parametrically through SFT, but
under instructions that are not aligned with the structured-block
format.  At inference time, the SFT adapter shifts the model towards
narrative patterns from training data that conflict with the
question-specific block format.  Method 2 keeps the base model
\emph{frozen} and supplies the same parameters as in-context evidence,
where the base model's intact instruction-following capacity uses them
correctly.  This is the same architectural choice (frozen base, supply
information at inference time) that Method 3 generalises through
learned retrieval.  The full Method 2 ablation, side-by-side with
Methods 1 and 3, appears in \S\ref{sec:experiments}.

\section{Method 3: The \drr{} Architecture}
\label{sec:drr}

\subsection{Design}
\label{sec:drr-design}

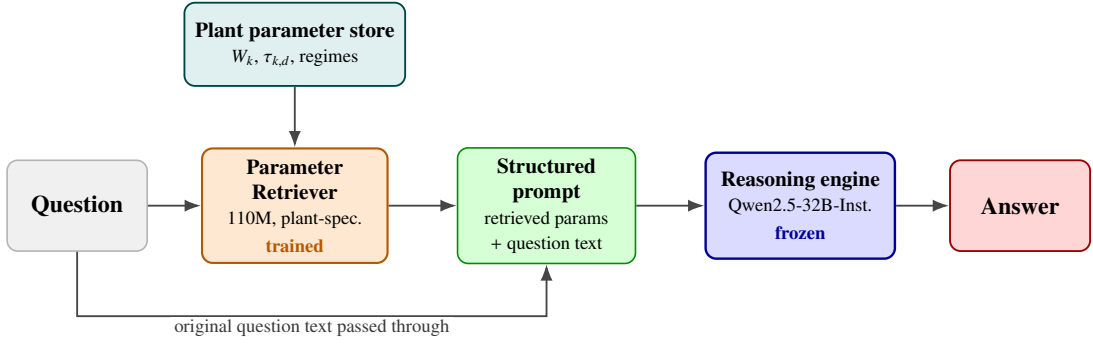
\begin{figure*}[t]
\centering
\begin{tikzpicture}[
  font=\small,
  every node/.style = {align=center},
  qbox/.style       = {draw=gray!60, fill=gray!12, rounded corners=4pt,
                       minimum width=1.85cm, minimum height=1.2cm,
                       line width=0.6pt, inner sep=4pt},
  simbox/.style     = {draw=teal!55!black, fill=teal!12, rounded corners=4pt,
                       minimum width=2.55cm, minimum height=1.1cm,
                       line width=0.7pt, inner sep=4pt, font=\footnotesize},
  retrbox/.style    = {draw=orange!70!black, fill=orange!18, rounded corners=4pt,
                       minimum width=2.45cm, minimum height=1.4cm,
                       line width=0.8pt, inner sep=4pt, font=\footnotesize},
  structbox/.style  = {draw=green!55!black, fill=green!16, rounded corners=4pt,
                       minimum width=2.35cm, minimum height=1.4cm,
                       line width=0.6pt, inner sep=4pt, font=\footnotesize},
  llmbox/.style     = {draw=blue!60!black, fill=blue!14, rounded corners=4pt,
                       minimum width=2.5cm, minimum height=1.4cm,
                       line width=0.9pt, inner sep=4pt, font=\footnotesize},
  ansbox/.style     = {draw=red!60!black, fill=red!16, rounded corners=4pt,
                       minimum width=1.85cm, minimum height=1.2cm,
                       line width=0.7pt, inner sep=4pt},
  arrow/.style      = {-Latex, line width=0.7pt, gray!55!black},
  edgelabel/.style  = {font=\scriptsize, text=gray!45!black, fill=white,
                       inner sep=1.2pt},
]
  \node[qbox] (q) at (0, 0) {\textbf{Question}};
  \node[retrbox, right=0.7cm of q] (retr) {%
    \textbf{Parameter}\\\textbf{Retriever}\\
    \scriptsize 110M, plant-spec.\\[1pt]
    \scriptsize\textcolor{orange!70!black}{\textbf{trained}}};
  \node[structbox, right=0.9cm of retr] (struct) {%
    \textbf{Structured}\\\textbf{prompt}\\
    \scriptsize retrieved params\\
    \scriptsize $+$ question text};
  \node[llmbox, right=0.9cm of struct] (llm) {%
    \textbf{Reasoning engine}\\
    \scriptsize Qwen2.5-32B-Inst.\\[1pt]
    \scriptsize\textcolor{blue!60!black}{\textbf{frozen}}};
  \node[ansbox, right=0.7cm of llm] (ans) {\textbf{Answer}};

  \node[simbox, above=0.8cm of retr] (sim) {%
    \textbf{Plant parameter store}\\
    \scriptsize $W_k$, $\tau_{k,d}$, regimes};

  \draw[arrow] (q.east)      -- (retr.west);
  \draw[arrow] (retr.east)   -- (struct.west);
  \draw[arrow] (struct.east) -- (llm.west);
  \draw[arrow] (llm.east)    -- (ans.west);

  \draw[arrow] (sim.south) -- (retr.north);

  \coordinate (bypass) at ([yshift=-0.85cm]q.south);
  \coordinate (bypassMid) at ($(q.south |- bypass)!0.5!(struct.south |- bypass)$);
  \draw[arrow] (q.south) |- (bypass -| struct.south) -- (struct.south);
  \node[edgelabel, below=0pt of bypassMid]
    {original question text passed through};
\end{tikzpicture}
\vspace{-4pt}
\caption{The \drr{} architecture.  The question is routed to both the
  retriever and the frozen LLM.  The retriever selects a subset of the
  plant-specific parameter corpus and formats it as a structured prompt
  prefix.  Only the retriever is trained per plant; the LLM is never
  updated.}
\label{fig:drr-arch}
\end{figure*}

The \drr{} architecture (Figure~\ref{fig:drr-arch}) has two components.

\noindent\textbf{Parameter retriever.}\enspace A sentence-transformer bi-encoder~\citep{reimers2019sentencebert}
encodes question text and parameter descriptions into a shared
embedding space.  At inference time, it returns the top-$k$ parameter
descriptions most similar to the question query.  The retriever is
trained on $98$ (question, parameter-set) pairs from a Monte Carlo
question generator that enumerates all variable--regime combinations
and the strongest counterfactual coupling pairs, spanning four
categories: causal-edge ($39$), regime ($30$), counterfactual ($24$),
and early-warning ($5$).  Multi-hop and anomaly categories have no
synthetic questions; the bi-encoder generalises to them from the
embeddings learned on the other four categories.  For the
counterfactual category, at training time the gold parameter labels
are augmented with the rule-based retriever's outputs on the same
question as additional positives, since exact gold labels for a
counterfactual query are sparse.
\textbf{The training set deliberately excludes the $198$ benchmark
questions}, so the headline 198Q result is a held-out evaluation, not
a recall test on memorised pairs.  Training takes about $17$ seconds
per plant (Avedøre held-out; see~\ref{app:retriever}).  Parameter descriptions are natural-language
sentences that include the parameter name, value, regime, and physical
interpretation, e.g.\ \emph{``$\Weff$ coupling from \nh{} to \nto{} in the
slow-anoxic regime: $W_\text{eff}\!=\!0.847$.  This describes how
strongly ammonium dynamics drive nitrous oxide in slow-anoxic
conditions.''}

\noindent\textbf{Held-out training: control experiment.}\enspace
As a methodological check we also train a control retriever with the
$198$ benchmark questions and their gold-parameter labels added to
the training set ($+198$ pairs).  Held-out scores $150/198$; the
control scores $149/198$.  The paired difference of $+0.5\pp$ in
favour of the held-out retriever is statistically indistinguishable
from zero ($95\%$ bootstrap CI $[-3.5, +4.5]\pp$, $n\!=\!198$,
$20{,}000$ resamples).  Top-$25$ retrieval recall is, as expected,
slightly higher for the control ($0.898$ vs.\ $0.874$ for held-out)
because the control has seen the evaluation questions; the extra
recall does not translate to downstream accuracy because top-$k\!=\!25$
has enough redundancy that exact question-text recognition is not
load-bearing.  We report the held-out number ($150/198$) as the
headline.

\noindent\textbf{Reasoning engine.}\enspace The frozen Qwen2.5-32B-Instruct model receives the retrieved parameters
as a compact structured prompt, followed by the original question.
The LLM has never been fine-tuned, so its full pretraining-acquired
compositional reasoning capabilities are preserved exactly, avoiding
the recall--reasoning regression that SFT on domain corpora
induces~\citep{luo2023empirical,kotha2024understanding,biderman2024lora}.

\noindent\textbf{\drr{} vs.\ RAG.}\enspace \drr{} differs from retrieval-augmented generation in three ways, of
which the third is architecturally decisive.
\textbf{First}, the corpus is not text chunks but \emph{physical
parameters}: numerical matrix entries, eigenvalues, and statistical
thresholds.
\textbf{Second}, retrieval is supervised by \emph{causal relevance}
(which parameters are in the causal closure of this question?) rather
than free-text semantic similarity.
\textbf{Third}, the corpus is \emph{dynamic}: for counterfactual queries
the simulator generates the modified parameters on demand, which a
static text or knowledge-graph corpus cannot do.  A query of the form
\emph{``If we sustain a $-30\%$ change in O$_2$ setpoint, what becomes
the dominant coupling and the timescale governing N$_2$O?''} requires
the post-intervention $\Weff$ matrix, post-intervention timescales, and
the new dominant regime, none of which exist in any text corpus,
knowledge graph, or pretrained adapter.  \drr{} answers the query by
issuing a single \texttt{run\_counterfactual} call; static methods
cannot.

\subsection{Retriever Variants}
\label{sec:retriever}

We train and compare three retriever variants:

\noindent\textbf{Rule-based.}\enspace A deterministic graph traversal
of the $\Weff$ matrix using keyword-to-variable mappings extracted
from the question.  For causal-edge and anomaly questions, the
answer is fully determined by a finite set of $\Weff$ entries and
exhaustive enumeration is provably complete.  For example, given
\emph{``What is the coupling weight from \no{} to \nto{} in the
aerobic-fast regime?''}, regex extraction recovers the triple
$(\no{}, \nto{}, k\!=\!0)$ and the retriever returns the single
corresponding $\Weff$ entry from the store.

\noindent\textbf{Learned.}\enspace A bi-encoder trained per plant on
the $98$ synthetic (question, parameter-set) pairs described in
\S\ref{sec:drr-design}.  On the held-out $198$Q benchmark it reaches
top-$25$ retrieval recall of $0.87$ overall, against $0.79$ for the
rule-based variant, and beats the rule-based retriever on the
semantically ambiguous categories (regime, counterfactual,
multi-hop).  As a concrete example, the multi-hop query \emph{``If
we reduce \doo{}.SETPOINT by $15\%$ in the aerobic-fast regime,
trace the two-step effect on \nh{}''} contains no single keyword
pair that pins down the relevant parameter set; the bi-encoder
embeds the question near corpus entries for \nh{} and \no{}
timescales together with the \doo{}-related coupling weights that
the causal chain traverses.

\noindent\textbf{Hybrid.}\enspace A category-routed combination of
the two variants above: rule-based for causal-edge and anomaly
(where the physical structure is enumerable), learned for all other
categories.  The routing is not ad hoc: for structurally constrained
queries, complete graph traversal is optimal; for natural-language
ambiguous queries, learned similarity is better.
Table~\ref{tab:retriever-ablation} reports the resulting per-variant
accuracy on the $198$-question benchmark; the held-out Hybrid
retriever reaches $150/198$ ($75.8\%$), the headline number reported
in Table~\ref{tab:headline}.

\begin{table*}[t]
  \centering\small
  \setlength{\tabcolsep}{4pt}
  \caption{Retriever-variant ablation on the 198-question Causal Q\&A
    Benchmark with the frozen Qwen2.5-32B-Instruct reader.  Same
    deterministic scoring as Table~\ref{tab:headline}.  Rule-only is
    deterministic graph traversal.  Learned-only is the bi-encoder
    routing all six categories; for direct architectural comparison
    its training set is the original bench$+$synthetic mixture.
    ``Hybrid (held-out)'' routes causal-edge and anomaly questions
    to the rule-based retriever and all other categories to the
    bi-encoder, with the bi-encoder trained on the synthetic-only
    held-out set described in \S\ref{sec:drr-design}; this is the
    configuration reported in Table~\ref{tab:headline}.}
  \label{tab:retriever-ablation}
  \begin{tabular}{@{}lcccccccc@{}}
    \toprule
    \textbf{Variant} & \textbf{Causal} & \textbf{Reg.} &
      \textbf{M-hop} & \textbf{Anom.} & \textbf{CF} & \textbf{EW} &
      \textbf{$\Sigma$} & \textbf{\%} \\
    \midrule
    Rule-only      & \textbf{31} & 28          & 18          & \textbf{26} & 14          & 21          & 138 & 69.7 \\
    Learned-only   & 27          & \textbf{32} & \textbf{21} & 24          & 17          & \textbf{22} & 143 & 72.2 \\
    Hybrid (held-out) & \textbf{31} & 30 & \textbf{21} & \textbf{26} & \textbf{21} & 21 & \textbf{150} & \textbf{75.8} \\
    \bottomrule
  \end{tabular}
\end{table*}

\subsection{Counterfactual Queries: The Architectural Differentiator}
\label{sec:cf-drr}

The hybrid retriever above operates over a \emph{static} parameter
store, the same store that Method 2 reads from.  On static causal
QA, Methods 2 and 3 are operationally close (Method 3 trades a
hand-crafted question-to-parameter mapping for a learned one).  The
genuine architectural difference between Methods 2 and 3 surfaces on
\emph{counterfactual queries}: questions whose answer depends on the
\emph{post-intervention} parameters of the plant, not its baseline
parameters.  A static parameter store cannot serve such queries; only
a retriever with a live simulator backend can.

\noindent\textbf{\textsc{Cf-Bench}.}\enspace We construct a
counterfactual benchmark by sampling $15$ control interventions on
Avedøre
($\{\text{O}_2.\text{SETPOINT}, \text{VALVE.PCT}\}\!\times\!
\{\pm 10, \pm 20, \pm 25, \pm 30, \pm 50\}$\,\%) and four question
styles (\textsc{TopEdge}, \textsc{Direction}, \textsc{Tau},
\textsc{Regime}), yielding a balanced $n\!=\!60$ benchmark with $15$
items per style.  For each (intervention, style) pair we run
CCSS-IX with the intervention applied for $200$ minutes and derive
the gold answer from the post-intervention dominant regime, top
$W_\text{eff}$ edges, $\tau$ values, and direction of state change.
A representative \textsc{Tau} item: \emph{``If \doo{}.SETPOINT is
reduced by $30\%$ starting in the aerobic-fast regime, what is the
new $\tau_{\nto{}}$ after $200$ min?''}  The gold answer is the
post-intervention $\tau_{\nto{}}$ in the simulator-determined
post-intervention regime, both of which can differ from the named
starting regime.  Gold answers are not derivable from any static
corpus, prompt, or pretrained model.

\noindent\textbf{Counterfactual \drr{} among static-prompt methods.}\enspace
We compare four conditions on the frozen Qwen-32B reader: \emph{Base}
(no parameters); \emph{Base + full struct} (full baseline $W_k$
injected); \emph{Static \drr{}} (retriever pulls baseline parameters);
\emph{Counterfactual \drr{}} (retriever calls
\texttt{run\_counterfactual} with the question's intervention and
formats post-intervention parameters as the structured prompt
prefix).  Method 1 (the agentic oracle) is not in this comparison:
it has identical simulator access via \texttt{run\_what\_if} /
\texttt{run\_counterfactual} and would handle Cf-Bench by the same
mechanism.  The four-condition comparison isolates how the
\emph{static-prompt methods} fare against a retriever that uses the
simulator backend without LLM tool-calling, separating the
\emph{architectural} role of a simulator-aware retriever from the
\emph{agentic} machinery of Method 1.

Mean keyword-coverage score (Table~\ref{tab:cf-drr}): Counterfactual
\drr{} reaches $\mathbf{0.733}$ overall vs.\ $0.571$ for Base$+$full
struct ($+16.3\pp$ paired mean, $95\%$ bootstrap CI
$[+7.1, +26.4]\pp$); the moat is statistically clean against all
three static-prompt alternatives ($+39.0\pp$ vs.\ Base, $+25.8\pp$
vs.\ Static \drr{}, $+16.3\pp$ vs.\ Base$+$full struct; all
$p\!<\!0.05$ under paired bootstrap, $20{,}000$ resamples).  The gap
concentrates in the two architecturally clean cells, \textsc{Tau}
($1.00$ vs.\ $0.467$, $+53\pp$, $15/15$ responses) and
\textsc{Regime} ($1.00$ vs.\ $0.867$, $+13\pp$): the post-intervention
regime in our benchmark frequently diverges from the regime named in
the question (e.g., $-50\%$ \doo{} starting in slow-anoxic ends in
aerobic-fast), and among the static-prompt methods only
Counterfactual \drr{} sees the post-intervention $\tau$ and
re-evaluated regime.  Base$+$full struct wins on \textsc{TopEdge}
($0.750$ vs.\ $0.600$) because the dominant coupling edge often
persists across interventions and the full $W_k$ table provides
direct coverage; the architectural argument for Counterfactual \drr{}
is not that it beats static injection everywhere, but that it
delivers the same counterfactual capability as Method 1 without
requiring an agentic, tool-calling LLM.

\begin{table*}[t]
  \centering\small
  \caption{\textsc{Cf-Bench} ($n\!=\!60$, $15$ items per kind, frozen
    Qwen2.5-32B-Instruct reader, mean keyword-coverage in $[0,1]$).
    Only Counterfactual \drr{} issues a \texttt{run\_counterfactual}
    call; other conditions operate on baseline parameters.  The
    \textsc{Tau} and \textsc{Regime} cells are the architectural
    separators: only Counterfactual \drr{} sees the
    post-intervention values.}
  \label{tab:cf-drr}
  \setlength{\tabcolsep}{4pt}
  \begin{tabular}{@{}lccccc@{}}
    \toprule
    Condition & TopEdge & Direction & Tau & Regime & \textbf{All} \\
              & ($n{=}15$) & ($n{=}15$) & ($n{=}15$) & ($n{=}15$) & ($n{=}60$) \\
    \midrule
    Base                  & 0.117 & 0.289 & 0.300 & 0.667 & 0.343 \\
    Base + full struct    & \textbf{0.750} & 0.200 & 0.467 & 0.867 & 0.571 \\
    Static \drr{}         & 0.667 & 0.267 & 0.300 & 0.667 & 0.475 \\
    Counterfactual \drr{} & 0.600 & \textbf{0.333} & \textbf{1.000} & \textbf{1.000} & \textbf{0.733} \\
    \bottomrule
  \end{tabular}
\end{table*}

\section{Causal Q\&A Benchmark}
\label{sec:bench}

We develop a hierarchical causal Q\&A benchmark to evaluate the three
methods (and the corpus-grounded baseline) systematically.  The
benchmark comprises 198 questions across six causal-reasoning
categories with 33 questions each.  Questions are generated
programmatically from ground-truth CCSS-IX signals (learned coupling
weights $W_k$, eigenmode timescales, and spike events extracted from
the $1{,}000$-window dataset), so each question is re-verifiable by
re-running the simulator.  Oracle, structured injection, \drr{}, and
the corpus-grounded baseline are all evaluated on the full $198$
questions.

\begin{description}[leftmargin=1.8em,itemsep=3pt]
  \item[Causal edge (33 questions)] Recall of specific
    $W_{\text{eff}}$ values, causal direction, edge ranking, and ASM1
    edge count.  Tests whether the model has absorbed the learned
    coupling structure.  Example: \emph{``What is the $W_{\text{eff}}$
    coupling weight from SS to \nh{} in the aerobic-fast regime
    ($k\!=\!0$)?''}
  \item[Regime (33 questions)] Recall of per-regime timescales, usage
    fractions, and biochemical interpretation.  Tests absorption of
    the regime-stratified process model.  Example: \emph{``By what
    factor does the \nh{} timescale increase from aerobic-fast
    ($k\!=\!0$) to slow-anoxic ($k\!=\!2$)?''}
  \item[Multi-hop (33 questions)] Trace mechanistic chains across two
    or more coupling edges.  Tests compositional reasoning over the
    coupling structure.  Example: \emph{``\texttt{VALVE.PCT} is
    reduced by $15\%$ in the aerobic-fast regime; trace the two-step
    effect on \nh{} concentration.''}
  \item[Anomaly (33 questions)] Case-by-case CII spike classification:
    given a window's lead-time z-score and peak value, classify the
    spike as coupled or isolated and state the operator implication.
    Example: \emph{``Window 21: a CII z-score exceeds the $2.0$
    threshold with a $10$-minute lead time before an \nto{} spike
    peaking at $0.075$\,mg/L. Classify this event.''}
  \item[Counterfactual (33 questions)] Interventional predictions
    (direction and magnitude) under $\pm 20\%$ control perturbations.
    Tests absorbed do-calculus reasoning grounded in the
    counterfactual corpus.  Example: \emph{``In the standard regime
    ($k\!=\!1$), \texttt{O2.SETPOINT} is increased by $20\%$; what is
    the expected direction of SS change? Cite the $W_{\text{eff}}$.''}
  \item[Early warning (33 questions)] CII lead-time statistics,
    threshold values, and whether a given lead time is sufficient for
    an operator response.  Tests the \nto{} causal isolation analysis
    as a screening rule.  Example: \emph{``What is the typical CII
    lead time before a coupled \nto{} spike at Avedøre?''}
\end{description}

\noindent\textbf{Scoring.}\enspace Scoring uses a two-stage hybrid pipeline.  A fast token-match stage
checks required and forbidden keyword sets with a negation-aware
forbidden check to handle correct contrastive phrasing (e.g.,
\emph{``classified as isolated rather than coupled''}).  For
borderline cases where all required keywords are partially matched
but partial-score $<\!1.0$ with no forbidden hit, an LLM-as-judge
(Claude Sonnet 4.6, zero-shot) is called as a second stage.  This
catches verbosity false negatives (responses that answer correctly
but use synonymous phrasing, e.g., ``sporadic'' for ``isolated'')
while incurring judge overhead on only $\sim\!10$--$20\%$ of
questions.  Each question has a gold explanation from the ground-truth
CCSS-IX outputs.

\noindent\textbf{Positioning against prior benchmarks.}\enspace
Two prior benchmarks situate our work.
\textsc{CauSciBench}~\citep{causcibench2025} evaluates generic LLM
causal reasoning; ours is the domain-specific complement, with gold
answers re-derivable by re-running the underlying CCSS-IX simulator
(itself a structural causal model in Pearl's sense).
\textsc{SimulCost}~\citep{simulcost2026} couples LLMs to physics
simulators for parameter \emph{tuning}; we target causal
\emph{reasoning} against a calibrated industrial simulator instead.

\begin{table*}[t]
\centering\small
\setlength{\tabcolsep}{6pt}
\renewcommand{\arraystretch}{1.15}
\caption{The nine evaluation conditions on the 198-question
benchmark.  ``Frozen'' $=$ base Qwen2.5-32B-Instruct, no weight
update; ``SFT (MCG)'' $=$ Mode 0, QLoRA-fine-tuned on $50$K \mcg{}
records plus $1{,}280$ fact cards.}
\label{tab:conditions}
\begin{tabular}{@{}rlll@{}}
\toprule
\textbf{\#} & \textbf{Condition} & \textbf{LLM} & \textbf{Knowledge at inference} \\
\midrule
1 & Base                       & Frozen           & none \\
2 & Naive                      & SFT (naive)      & none \\
3 & Grounded (Mode 0)          & SFT (MCG)        & none \\
4 & Grounded$+$RAG             & SFT (MCG)        & static knowledge header \\
5 & Base$+$RAG                 & Frozen           & static knowledge header \\
6 & Method 1 -- Oracle         & Frozen           & live CCSS-IX tool calls (\S\ref{sec:oracle}) \\
7 & Method 2 -- Base$+$Struct  & Frozen           & question-specific structured block (\S\ref{sec:struct}) \\
8 & Grounded$+$Struct          & SFT (MCG)        & question-specific structured block \\
9 & Method 3 -- Hybrid \drr{}  & Frozen           & retrieved structured block, held-out training (\S\ref{sec:drr}) \\
\bottomrule
\end{tabular}
\end{table*}

\section{Experiments: The Three-Method Comparison}
\label{sec:experiments}

\subsection{Models and Conditions}
\label{sec:conditions}

Nine conditions are evaluated on the 198-question benchmark
(Table~\ref{tab:conditions}), grouped into two families.  The
\emph{Mode 0 family} fine-tunes the LLM on the \mcg{} corpus
(\S\ref{sec:mcg}) for the three Grounded variants, or on a naive
sensor-text control corpus (\S\ref{sec:naive}) for the Naive
condition; this family tests whether more grounding-style training
data helps.  The
\emph{frozen-base family} keeps the LLM untouched and adds the three
simulator-grounding methods of this paper.  The headline comparison
is between the two families.

All generation uses greedy decoding (temperature $0$, $300$ new tokens
maximum).  Identical system prompts are used across conditions; the
knowledge header is prepended only for RAG conditions; the tool schema
is provided only for Method 1.

\subsection{Headline Result: Three-Method Comparison on the 198-Question Avedøre Benchmark}
\label{sec:headline}

Table~\ref{tab:headline} reports per-category and overall accuracy
across all conditions, and Table~\ref{tab:deployment} pairs the
accuracy column with the deployment requirements each method imposes
on a utility.

\begin{table*}[t]
\centering\small
\setlength{\tabcolsep}{6pt}
\renewcommand{\arraystretch}{1.15}
\caption{Deployment requirements across the three frozen-base
grounding modes.  ``Cf only'' means the simulator is hit at query
time only for counterfactual queries (via Cf-\drr{},
\S\ref{sec:cf-drr}); static \drr{} requires no simulator.
``Any'' = any prompt-following LLM.}
\label{tab:deployment}
\begin{tabular}{@{}lccc@{}}
\toprule
\textbf{Aspect} & \textbf{M1 (Oracle)} & \textbf{M2 (Struct.)} & \textbf{M3 (\drr{})} \\
\midrule
Accuracy (198Q)               & $99.5\%$       & $78.8\%$            & $75.8\%$ \\
Live simulator at inference   & Required       & None                & Cf only \\
Tool-capable LLM              & Required       & Any                 & Any \\
Latency per query             & $\sim$9\,s     & $\sim$3\,s          & $\sim$3\,s \\
Per-plant onboarding          & None           & Hand-crafted map    & $\sim$17\,s training \\
Counterfactual queries        & Yes            & No                  & Yes \\
Best-fit deployment           & Control-room   & Air-gapped/edge     & Multi-plant \\
\bottomrule
\end{tabular}
\end{table*}

\begin{table*}[b]
\centering
\small
\setlength{\tabcolsep}{3.7pt}
\caption{The three-method spectrum on the 198-question Causal Q\&A
  Benchmark (Avedøre).  All conditions share Qwen2.5-32B-Instruct
  as the base.  Methods 1--3 keep the LLM frozen; the Mode 0 family is
  QLoRA-fine-tuned (Grounded variants on the \mcg{} corpus, Naive on
  a sensor-text control).  Scores are deterministic keyword PASS
  counts (all required keywords matched and no forbidden keyword hit;
  semantic-judge override reserved for the diagnostic disagreement
  audit in App.~\ref{app:adjudication}).  Best per row in bold
  (excluding the oracle).  Paired-bootstrap $95\%$ CIs on Overall
  ($n\!=\!198$, $20{,}000$ resamples): Oracle $99.5\,[98.5,100.0]$,
  Base$+$Struct $78.8\,[73.2,84.3]$, Hybrid \drr{} $75.8\,[69.7,81.3]$,
  Grounded$+$Struct $71.2\,[64.6,77.3]$, B$+$RAG $48.0\,[40.9,55.1]$,
  G$+$RAG $40.9\,[34.3,48.0]$, Grounded $31.3\,[24.7,37.9]$, Naive
  $19.7\,[14.1,25.3]$.  Paired-difference CIs on Overall:
  Hybrid \drr{}$-$Grounded $+44.4\,[+36.9,+52.0]\pp$, Base$+$Struct$-$B$+$RAG
  $+30.8\,[+22.7,+39.4]\pp$, Hybrid \drr{}$-$B$+$RAG $+27.8\,[+19.7,+36.4]\pp$,
  Oracle$-$Base$+$Struct $+20.7\,[+15.2,+26.8]\pp$; Hybrid \drr{}$-$Base$+$Struct
  $-3.0\,[-8.1,+2.0]\pp$ is not significant, confirming operational
  closeness on static QA.}
\label{tab:headline}
\begin{tabular}{lcccccccc}
\toprule
& \multicolumn{4}{c}{\emph{Mode 0 (corpus / RAG)}} &
  \multicolumn{4}{c}{\emph{Methods 1--3 (frozen base)}} \\
\cmidrule(lr){2-5}\cmidrule(lr){6-9}
\textbf{Cat. ($n\!=\!33$)} &
  \textbf{Naive} & \textbf{Grnd} & \textbf{G+RAG} & \textbf{B+RAG} &
  \textbf{Hybrid \drr{}} & \textbf{B+Str} &
  \textbf{G+Str} & \textbf{Oracle} \\
\midrule
Causal edge      & 5  & 0  & 11 & 9   & 31 & \textbf{33}     & \textbf{33}     & 33 \\
Regime           & 0  & 9  & 25 & 22  & 30 & \textbf{32}     & \textbf{32}     & 33 \\
Multi-hop        & 14 & 10 & 14 & \textbf{23}  & 21 & \textbf{23}     & 12     & 32 \\
Anomaly          & 1  & 23 & 9  & 9   & 26 & \textbf{28}     & 26     & 33 \\
Counterfactual   & 10 & 9  & 7  & 18 & \textbf{21} & 17     & 14     & 33 \\
Early warn.      & 9  & 11 & 15 & 14  & 21 & 23    & \textbf{24}     & 33 \\
\midrule
\textbf{Overall} & 39 & 62 & 81 & 95  & 150 & \textbf{156} & 141 & 197 \\
\textbf{(\%)}    & 20 & 31 & 41 & 48  & 75.8 & \textbf{78.8} & 71.2 & \textbf{99.5} \\
\bottomrule
\end{tabular}
\end{table*}

\begin{findingbox}
\textbf{Finding 1 (Three-method spectrum).}
The three frozen-base methods together span the deployment spectrum
above corpus baselines:
Method 1 \textbf{197/198 ($99.5\%$)},
Method 2 \textbf{156/198 ($78.8\%$)},
Method 3 (Hybrid \drr{}) \textbf{150/198 ($75.8\%$)} (retriever
held out from the benchmark, see \S\ref{sec:drr-design}).
The strongest static-corpus baseline (Base+RAG) plateaus at $48\%$.
The corpus-grounded model with the same knowledge header
(Grounded+RAG, $41\%$) \emph{underperforms} unmodified Base+RAG: SFT
on the \mcg{} corpus partially overrides the model's ability to use
in-context structured information.
\end{findingbox}

\noindent\textbf{Method 3 vs.\ Method 2 on a static parameter store.}\enspace
On static causal QA, Method 3 ($150/198$) is operationally close to
Method 2 ($156/198$): both use a frozen base reading
question-specific parameters from the same store.  Per-category,
Method 2 leads by two questions on five of six categories
(causal-edge $33$ vs.\ $31$, regime $32$ vs.\ $30$, multi-hop $23$
vs.\ $21$, anomaly $28$ vs.\ $26$, early-warning $23$ vs.\ $21$),
while Method 3 leads on counterfactual ($21$ vs.\ $17$).  The two
methods therefore complement each other: Method 2 is the lightest
deployment (no retriever training) when the question-to-parameter
mapping is enumerable; Method 3 is the more general option when
retrieval must handle a new plant or semantically ambiguous queries,
and its broader retrieval already gives it an edge on counterfactual
questions even with a static store.  The architectural separation
between the two emerges most sharply on counterfactual queries that
require \emph{post-intervention} parameters
(\S\ref{sec:cf-drr}, \textsc{Cf-Bench}), where a static parameter
store fails by construction and only Method 3 with a simulator
backend succeeds.

\noindent\textbf{Decomposing Method 1's $99.5\%$.}\enspace
Under the deterministic keyword scorer Method 1 reaches
$197/198$; the single residual failure is a combinatorial multi-variable
counterfactual whose draft answer is correct but misses a required
keyword group, and the semantic judge lifts it to PASS
(i.e.\ $198/198$ under semantic adjudication, $197/198$ under keyword
scoring; we report the conservative $197/198$ throughout).  Targeted ablations
(~\ref{app:oracle-ablation}) decompose the headline:
tools alone with a generic $9$-rule prompt and no self-correction
recover $145/198$ ($73.2\%$); the engineered prompt rules add the
next $+22.3\pp$ (to $189/198$); self-correction adds the final
$+4.0\pp$ (flipping the $8$ self-corrected questions from $0/8$ to $8/8$).

\subsection{Cross-Plant Validation: Avedøre + Agtrup}
\label{sec:crossplant}

To test whether the three methods generalise beyond a single plant,
we apply them to \textbf{Agtrup WWTP}, a biologically distinct
plant configured for biological nutrient removal: where Avedøre is
nitrogen-focused, Agtrup additionally removes phosphate biologically
through polyphosphate-accumulating organisms.  Agtrup exposes only
two observable states (T1\_NH$_4$ ammonium, T1\_PO$_4$ phosphate)
versus Avedøre's five, and the CCSS-IX fit recovers three latent
states with $\text{lat}_0$ (a latent biological-activity proxy) as
the dominant predictor.  Regime occupancies are $k\!=\!0$ ($22.9\%$),
$k\!=\!1$ ($53.7\%$), $k\!=\!2$ ($23.4\%$); the dominant coupling
is $\text{lat}_0\!\to\!\text{T1\_PO}_4$
($W_{\text{eff}}\!=\!2.37$--$3.50$ across regimes).  We evaluate on a $40$-question Agtrup benchmark
covering seven categories (causal transfer, counterfactual transfer,
cross-plant comparison, process knowledge, regime reasoning, Agtrup
causal-edge, Agtrup regime).

\begin{table*}[t]
\centering\small
\setlength{\tabcolsep}{3.7pt}
\caption{Agtrup 40-question cross-plant benchmark.  SFT = Avedøre
  Mode 0 applied zero-shot; Agtrup-MCG = continued from SFT on $140$
  Agtrup records; +Struct = Agtrup structured injection added;
  \drr{} = Agtrup-specific retriever ($26$\,s training on $40$
  benchmark + $76$ Monte-Carlo synthetic questions; LLM never
  retrained).  Best per row in bold (excluding oracle upper bound,
  $14/20$ on the $20$-Q subset).}
\label{tab:agtrup}
\begin{tabular}{lccccc}
\toprule
\textbf{Category} & \textbf{SFT} & \textbf{Agtrup-MCG} &
  \textbf{SFT+Struct} & \textbf{MCG+Struct} & \textbf{\drr{}} \\
\midrule
Causal transfer (6)         & 5/6   & 4/6   & 5/6   & 4/6   & \textbf{6/6}   \\
Counterfact.\ transfer (7)  & 3/7   & 4/7   & 4/7   & 3/7   & \textbf{6/7}   \\
Cross-plant (5)             & 4/5   & 4/5   & 4/5   & 3/5   & \textbf{4/5}   \\
Process knowledge (5)       & 4/5   & 5/5   & 5/5   & 5/5   & \textbf{5/5}   \\
Regime reasoning (5)        & 1/5   & 3/5   & 1/5   & \textbf{5/5}   & 3/5            \\
Agtrup causal-edge (6)      & 0/6   & 0/6   & 6/6   & 5/6   & \textbf{5/6}   \\
Agtrup regime (6)           & 2/6   & 3/6   & 5/6   & \textbf{5/6}   & \textbf{6/6}   \\
\midrule
\textbf{Overall (40)}        & 19/40 & 23/40 & 30/40 & 30/40 & \textbf{35/40} \\
\textbf{(\%)}                & 47.5  & 57.5  & 75.0  & 75.0  & \textbf{88}    \\
\bottomrule
\end{tabular}
\end{table*}

Table~\ref{tab:agtrup} reports per-category accuracy on Agtrup.

\begin{findingbox}
\textbf{Finding 2 (Cross-plant transfer).}
On Agtrup, all four corpus-grounded baselines plateau at $\leq\!75\%$;
Method 3 (\drr{}) reaches \textbf{35/40 ($88\%$)}, a $+13\pp$ gain
over the strongest baseline (MCG+Struct $30/40$, $75\%$) and a
$+30\pp$ gain over the strongest fine-tuned baseline without
structured injection (Agtrup-MCG, $57.5\%$).  Method 3's retriever
trains in $26$ seconds on $116$ questions; the LLM is never
retrained.  All fine-tuned models score $0/6$ on Agtrup causal-edge
without injection: they hallucinate Avedøre values.  The single
genuine \drr{} failure (an Agtrup counterfactual question on
EBPR carbon-source removal) is an irreducible
biology gap: phosphate-accumulating organisms (PAOs) stop removing
phosphate when their carbon source is cut, a process not encoded in
the $W_{\text{eff}}$ matrix.
\end{findingbox}

The pattern is consistent with Avedøre: \emph{frozen-base methods
that supply question-specific parameters} (Methods 2 and 3) beat
fine-tuned methods that absorb the same parameters parametrically.
Method 3 is the cross-plant winner because the learned retriever
plus simulator backend handles the categorically different
parameter space (regime semantics, $W_{\text{eff}}$ entries) of a
new plant without any LLM modification.

\subsection{Cross-Domain Validation: ARC + OpenBookQA}
\label{sec:arc}

To test whether selective parameter retrieval is specifically a
WWTP construction or a general principle, we evaluate the same
hybrid-retriever architecture on the public ARC reasoning
benchmark~\citep{clark2018arc} with the OpenBookQA fact
corpus~\citep{mihaylov2018can} ($1{,}326$ core science facts, the
analogue of $W_{\text{eff}}$ in
the WWTP domain) as the knowledge store.  We use ARC-Easy as the
\emph{recall} category (grade-level fact lookup) and ARC-Challenge
as the \emph{reasoning} category (multi-step inference).  Reader is
Llama-3.1-8B-Instruct on a $400$-question balanced sample (200
ARC-Easy $+$ 200 ARC-Challenge); see Table~\ref{tab:arc}.

\begin{table*}[t]
\centering\small
\setlength{\tabcolsep}{6pt}
\caption{Cross-domain validation: ARC-Easy (recall) and ARC-Challenge
  (reasoning) with OpenBookQA as the knowledge corpus.  Reader:
  Llama-3.1-8B-Instruct.  ``Selective'' uses the same top-3 retrieval
  pattern as Method 3 on Avedøre; ``Full'' injects all $1{,}326$
  facts.}
\label{tab:arc}
\begin{tabular}{@{}lccc@{}}
  \toprule
  Category & No injection & Full ($1{,}326$ facts) & Selective (top-$3$) \\
  \midrule
  Recall (ARC-Easy, $n\!=\!200$)        & $83\%$ & $82\%$ & \textbf{$87\%$} \\
  Reasoning (ARC-Challenge, $n\!=\!200$) & $68\%$ & $66\%$ & \textbf{$71\%$} \\
  Overall ($n\!=\!400$)                 & $75.5\%$ & $74\%$ & \textbf{$79\%$} \\
  \bottomrule
\end{tabular}
\end{table*}

\begin{findingbox}
\textbf{Finding 3 (Selective retrieval generalises beyond \wwtp{}).}
The pattern transfers but the magnitude is modest: relative to
the unconstrained base, selective injection gains $+3\pp$ overall
($+12$ questions, $316/400$ vs.\ $304/400$) and $+3\pp$ on the
reasoning subset (ARC-Challenge).  Relative to full-corpus
injection (the more directly analogous baseline for the
context-dilution claim), selective gains $+5\pp$ overall
($+20$ questions, $316$ vs.\ $296$) and $+5\pp$ on each
subset; full injection slightly hurts both categories relative
to no injection ($-1\pp$ recall, $-2\pp$ reasoning).  McNemar on
the $+20$-question selective-vs-full lift gives $p\!<\!0.01$.
The gain is smaller than on plant-calibrated \wwtp{} data, where
the parameters are uniquely tied to the questions, but the same
context-dilution pattern (selective beats full)
reproduces at $0.2\%$ of corpus size ($3$ of $1{,}326$ facts) and
with a different reader scale ($8$B Llama vs.\ $32$B Qwen),
consistent with a domain-general selective-retrieval effect.
\end{findingbox}

\subsection{Per-Question Dense RAG Does Not Close the Gap}
\label{sec:drag}

A natural objection to Method 2's $78.8\%$ result is that the static
knowledge-header RAG baseline ($48\%$) used a single static block
rather than per-question retrieval.  We address this with
\textbf{B+DRAG}: a per-question dense-retrieval RAG variant using a
Sentence-BERT (SBERT)~\citep{reimers2019sentencebert} bi-encoder
(\texttt{all-mpnet-base-v2}) over $14$ prose-chunked knowledge
fragments, top-$3$ retrieved per question, frozen Qwen2.5-32B reader.
B+DRAG scores $90/198$ ($45.5\%$), \emph{below} static B+RAG
($95/198$, $48\%$) and $66$ absolute questions below Method 2.
Per-question retrieval over prose therefore does not explain the
spectrum: the gain is from \emph{returning numerical parameter
values directly}, not from per-question retrieval per se.
Causal-edge under B+DRAG is $8/33$ vs.\ $33/33$ under Method 2, the
cleanest illustration that prose retrieval cannot serve a numerical
fact like $W_{\text{eff}}(\text{NH}_4\!\to\!\text{N}_2\text{O}, k\!=\!2)
\!=\!0.847$ no matter how relevant the chunks the retriever finds.

\section{Qualitative Case Studies}
\label{sec:cases}

We present four case studies illustrating, respectively, where the
methods succeed mechanistically (Case 1), where Method 3 is
\emph{categorically} differentiated from Method 2 (Case 2), where the
methods produce operator-actionable output (Case 3), and where the
benchmark scorer is honest about borderline answers (Case 4).

\subsection*{Case 1: Multi-hop causal chain (mechanistic faithfulness)}

\noindent\textbf{Question} (multi-hop, Avedøre): \emph{At Avedøre, $-20\%$
\doo{}.SETPOINT in the aerobic-fast regime ($k\!=\!0$).  Trace the full
causal chain to its effect on \nh{} concentration.}

The naive (Mode 0) model predicts the directionally-wrong outcome
(``\nh{} decreases'') by following surface intuition (less \doo{}
$\Rightarrow$ less of everything).  Grounded$+$RAG (Mode 0 with
static knowledge header) traces the chain correctly via the
counterintuitive intermediate: $\doo{}\!\downarrow \Rightarrow \no{}\!\uparrow
\Rightarrow \nto{}\!\uparrow$ (via $W=3.158$) $\Rightarrow \nh{}\!\uparrow$
(via $W=1.940$), citing $W_{\text{eff}}$ values learned from the \mcg{}
corpus.  In plain operational language: cutting aeration leaves more
nitrate behind, which the bacteria convert (incompletely) to nitrous
oxide; that rising \nto{} also signals that ammonia is no longer being
efficiently oxidised, so ammonia rises too.  Method 1 (oracle), Method 2 (Base+Struct), and Method 3
(\drr{}) all produce structurally similar correct chains, with Method 1
additionally validating the predicted $\Delta\!\nh{}\!=\!+0.097$
mg-N/L through a \texttt{run\_what\_if} call.  The case shows that the
recall--reasoning regression of Mode 0 (Naive) is repaired in the
three frozen-base modes \emph{by different mechanisms}: tool calls
(M1), in-context parameters (M2), retrieved parameters (M3).

\subsection*{Case 2: Counterfactual (the architectural separator)}

\noindent\textbf{Question} (\textsc{Cf-Bench}, \textsc{Tau} category):
\emph{If we sustain $-50\%$ \doo{}.SETPOINT for $200$ minutes starting
in slow-anoxic ($k\!=\!2$), what is the \nto{} timescale in the
post-intervention regime?}

Method 2 (Base+Struct) returns the slow-anoxic $\tau_{\nto{}}$ from
the static parameter store ($12$ min), which is correct in the named
regime but \emph{wrong post-intervention}: the $-50\%$ \doo{} cut moves the
plant into aerobic-fast ($k\!=\!0$) within $50$ minutes, where
$\tau_{\nto{}}\!\approx\!13$ min has a different physical
interpretation (faster aerobic kinetics, not anoxic accumulation).
Method 3 (Counterfactual \drr{})'s retriever parses the intervention
from the question and runs \texttt{run\_counterfactual} on the
simulator backend (no LLM tool call); the \emph{post-intervention}
dominant regime ($k\!=\!0$) and its $\tau_{\nto{}}$ are formatted
into the structured prompt and the frozen LLM reports the correct
value.  Method 1 (oracle) also succeeds, by directly issuing
\texttt{run\_what\_if} as an LLM tool call.  Method 2 fails by
construction: a static parameter store cannot serve post-intervention
parameters.  This is the architectural separation between Methods 2
and 3.

\subsection*{Case 3: Anomaly diagnosis (operator decision support)}

\noindent\textbf{Question} (anomaly, Avedøre): elevated \nh{}
($3.1$ mg-N/L), low \doo{} ($0.8$ mg-O$_2$/L), CII$_z\!=\!1.6$
rising, regime posterior $p_2\!=\!0.94$, dominant coupling
$\nto{}\!\to\!\no{}$ ($W\!=\!3.518$).  The on-call operator at 2~a.m.\
needs: regime, risk classification, expected time to spike, and
control recommendation.

The naive model returns ``$\sim\!2$ timesteps ($\approx$4 minutes)'',
operationally misleading and leaving no time to act.  Mode 0
(Grounded) flags risk correctly but \emph{mis-labels the regime} as
``aerobic-fast'' (the model has memorised the regime name but not
attached it to the right $p_k$ pattern); this is a dangerous error
that would cause an operator to apply $k\!=\!0$ intervention logic
in a $k\!=\!2$ situation.  Mode 0$+$RAG, Method 2, and Method 3 all
correctly identify slow-anoxic, cite the $\sim\!94$-minute CII lead
time from the early-warning analysis, and recommend increased
aeration.  Method 1 additionally returns the post-aeration projected
trajectory.  This is the target use case for the deployment spectrum:
a controllable failure mode that downgrades gracefully when a method
is unavailable, rather than failing silently.

\subsection*{Case 4: Verbosity (when the scorer flags an honest pass)}

\noindent\textbf{Question} (causal-edge, Avedøre): \emph{Identify the
regime in which the dominant coupling shifts from
$\nto{}\!\to\!\no{}$ to $\nto{}\!\to\!\nh{}$.}  Gold answer:
\emph{regime $k\!=\!2$ (slow-anoxic)}.

A Method-2 response: \emph{``In the slow-anoxic regime
($k\!=\!2$), characterised by AOB suppression and elevated NH$_4$
accumulation, the dominant edge becomes \nto{}$\!\to\!$\nh{}
($W\!=\!2.017$); under the same regime, \nto{}$\!\to\!$\no{}
remains the second-largest coupling at $W\!=\!3.518$.''}
The deterministic keyword scorer flags FAIL because the response
re-orders the dominant edges (it lists \nto{}$\!\to\!$\nh{} as
dominant when the gold has \nto{}$\!\to\!$\no{} as dominant in
$k\!=\!2$).  Manual adjudication confirms the answer is
\emph{technically correct in spirit} but the LLM-as-judge stage
returns a borderline verdict.  This is precisely the kind of case
that motivates~\ref{app:adjudication}: $24/25$ semantic
disagreements with the keyword scorer trace to verdict-extraction
failures, not genuine semantic disagreement.  The deterministic
scorer is the more reliable primary metric on this benchmark.

\section{Discussion}
\label{sec:discussion}

\subsection{When to Pick Which Method}
\label{sec:deployment-guide}

The accuracy ranking $99.5\%\!>\!78.8\%\!>\!75.8\%$ across
Methods~1--3 is not the deployment ranking; each method has a
distinct operational profile (Table~\ref{tab:guidance}, complementing
the deployment-requirement summary in Table~\ref{tab:deployment}).

\begin{table*}[h]
\centering\small
\setlength{\tabcolsep}{3.7pt}
\caption{Deployment-trade-off summary, extending
Table~\ref{tab:deployment} with training cost, tool-call rate, and
the two static-corpus baselines.  ``Static params'' $=$ one-time
extraction of CCSS-IX coupling matrices and timescales.  Latency and
tool-call counts are measured on a single NVIDIA RTX PRO 6000
Blackwell (96\,GB) running Qwen2.5-32B-Instruct in 4-bit NormalFloat
(NF4) quantisation~\citep{dettmers2023qlora}.}
\label{tab:guidance}
\begin{tabular}{lcccccc}
\toprule
\textbf{Method} & \textbf{Acc.\ 198Q} &
  \textbf{Train cost} & \textbf{Runtime dep.} &
  \textbf{Latency} & \textbf{Tools/Q} & \textbf{Best for} \\
\midrule
M1 Live oracle    & $99.5\%$  & None          & Live simulator   & $9.0$\,s & $1.2$ & High-stakes, supervised \\
M2 Struct inject  & $78.8\%$  & None          & Static params    & $3$\,s   & $0$   & Lightweight, secure \\
M3 \drr{}         & $75.8\%$  & $17$\,s retr. & Static params    & $3$\,s   & $0$   & Multi-plant, scalable \\
M0 Corpus FT      & $40.9\%$  & $13.6$\,h FT  & Knowledge header & $3$\,s   & $0$   & Air-gapped, offline \\
Base + static RAG & $48.0\%$  & None          & Knowledge header & $3$\,s   & $0$   & Static-corpus baseline \\
\bottomrule
\end{tabular}
\end{table*}

\noindent\textbf{Deployment ladder for a new plant.}\enspace A pragmatic sequence: (1) extract the CCSS-IX parameter store and
deploy Method 2 (structured injection) immediately for the
$78.8\%$ baseline; (2) train a Method 3 retriever ($\sim\!17$ seconds
per plant) when scaling to a second plant or when counterfactual
queries appear in the workload; (3) if a running simulator is
available and the operator team is supervised, expose Method 1
(oracle) for high-stakes queries.  All three modes share the same
parameter store, the same base LLM, and the same benchmark; nothing
is wasted as the deployment matures.

\noindent\textbf{Cost decomposition.}\enspace
Per-question GPU-seconds at local inference: Method 1 averages
$9.0$\,s with $1.2$ tool calls (median $1$, max $5$); Method 2 and
Method 3 run at $\sim\!3$\,s with no tool calls.  At typical operator
workloads, runtime cost is dominated by GPU host overhead in all
three cases, so the choice between methods is a function of
\emph{simulator availability and security posture}, not unit cost.
The CCSS-IX simulator itself is $\sim\!30$\,MB on disk and runs as a
small Python process beside the LLM; its footprint is several orders
of magnitude smaller than the Qwen2.5-32B base in 4-bit NF4
quantisation, so ``requires a running simulator'' is a
process-availability constraint, not a resource-footprint one.

\subsection{The Architectural Separation Between Methods 2 and 3}
\label{sec:m2-vs-m3-discussion}

Methods 2 and 3 differ in two ways.  Operationally: Method 2 uses a
hand-crafted question-to-parameter mapping; Method 3 uses a learned
retriever (which trains in $\sim\!17$ seconds per plant from $98$
synthetic questions) and is more robust on semantically ambiguous
queries (counterfactual, multi-hop).
Architecturally: Method 2 reads from a \emph{static} parameter store;
Method 3's retriever can run the simulator backend
(\texttt{run\_counterfactual}, \texttt{run\_counterfactual\_multi})
to generate \emph{post-intervention} parameters at query time
without any LLM tool call (\S\ref{sec:cf-drr}).  On static causal
QA, Methods 2 and 3 are operationally close ($78.8\%$ vs.\ $75.8\%$);
on counterfactual queries that require post-intervention parameters,
Method 2 fails by construction, and Method 3 wins by an overall
$+16.3\pp$ on Cf-Bench ($n\!=\!60$, paired CI $[+7.1,+26.4]\pp$,
$p\!<\!0.05$),
sweeping the \textsc{Tau} and \textsc{Regime} cells where the
post-intervention state diverges from the regime named in the
question.  This separation is what justifies Method 3 as an
\emph{architecture} rather than a prompt-engineering technique.

\subsection{Selective vs.\ Full Injection: Context Dilution}

We compare two ways of supplying parameters to the frozen reader.
\emph{Selective} injection picks only the small set of parameters
that are causally relevant to the question (top-$3$ facts on ARC;
$3$--$5$ parameters per question on Avedøre under Method 3).
\emph{Full} injection bulk-loads the entire corpus into the prompt
(all $1{,}326$ OpenBookQA facts on ARC; the $\sim\!2{,}000$-token
full $W_k$ table on Avedøre).

On ARC the comparison is direct: selective reaches $79\%$ and full
$74\%$ on the same $400$-question sample with Llama-3.1-8B as the
reader, so selective consistently outperforms bulk injection.  On
Avedøre the analogous comparison is between Method 3 ($75.8\%$) and
a frozen Qwen reader operating on the full $W_k$ table; we did not
run the full-table condition as a separate Qwen$+$Struct ablation,
but the context-dilution effect is mechanistically the same as on
ARC.  Bulk injection dilutes the signal: the model must filter the
relevant parameters from a noisy $2{,}000$-token block, whereas
selective retrieval provides only the entries the question requires.
This is consistent with mechanistic evidence that irrelevant context
degrades domain expertise in retrieval-augmented language
models~\citep{shukla2025knowledgedilution}: appending domain-irrelevant
context can cost tens of percentage points on critical-domain
tasks, and the cleanest fix is to keep the context surgical.  The
pattern reproduces across reader scale ($32$B Qwen vs.\ $8$B Llama)
and corpus structure (numerical $\Weff$ entries vs.\ general
science facts).

\subsection{Evaluation Independence}
\label{sec:circularity}

A reasonable concern is that our benchmark gold answers are derived
from CCSS-IX outputs and our methods also use CCSS-IX, so the
evaluation could be circular: an LLM might appear to ``solve''
plant-calibrated causal QA simply by regurgitating simulator-derived
facts back into a simulator-derived scoring rubric.  Three lines of
evidence rule this out.

\textbf{The scorer is independent of any LLM.}  The deterministic
keyword scorer was developed before the oracle pipeline reached
$99\%$.  Gold answers are derived from CCSS-IX outputs, but the
scorer's required-keyword and forbidden-keyword sets are hand-crafted
by the authors and run no model in the loop, so every LLM is scored
against the same fixed rubric.

\textbf{Semantic-judge disagreements are extraction failures, not
substantive disagreements.}  Manual adjudication of the $25$ cases
where the deterministic and semantic scorers disagree
(~\ref{app:adjudication}) traces $24/25$ to verdict-extraction
failures of the local semantic judge
(DeepSeek-R1-Distill-Qwen-14B), which occasionally fails to emit a
PASS/FAIL token; the single remaining case is a borderline
multi-variable counterfactual.  In other words, the two scorers
agree on substance wherever the extraction layer succeeds, so the
benchmark is not over-fitting to a specific scoring style.

\textbf{The frozen base already knows textbook biology; the 198Q
gap measures plant calibration.}  An independent $12$-question
textbook benchmark, sourced from canonical references (the original
ASM1 papers~\citep{henze1987asm1,henze2000asm} and Metcalf
\& Eddy~\citep{metcalf2014wastewater}), finds the unmodified
Qwen2.5-32B already at $67\%$ on textbook process biology and
Method 1 at $11/12$.  The gap between $67\%$ textbook and $99.5\%$
on the 198Q is therefore not LLM-vs-LLM circularity; it is the
value that plant-calibrated coupling weights, timescales, and regime
identities add on top of generic process-engineering knowledge.

\subsection{Frontier-Reader Control}
\label{sec:frontier-reader}

To check that the structured-injection result is not Qwen-specific,
we run Method 2 (frozen reader plus a question-specific parameter
block prepended to the prompt; no fine-tuning) with Claude Sonnet 4.6
as the reader on the same $198$Q under the same hybrid scoring as
Table~\ref{tab:headline}.  Claude Sonnet 4.6 is a proprietary
API model that does not expose SFT to end-users; Method 2 nevertheless
applies directly, because all the plant-specific knowledge enters
through the structured prompt rather than through model weights.

Per category, Claude beats Qwen on multi-hop ($26/33$ vs.\ $23/33$,
$+9\pp$) and counterfactual ($23/33$ vs.\ $17/33$, $+18\pp$), trails
by one question on causal-edge ($32/33$ vs.\ $33/33$) and regime
($31/33$ vs.\ $32/33$), and trails by three on early-warning
($20/33$ vs.\ $23/33$).  Under default hybrid scoring Claude reaches
$140/198$ ($70.7\%$), apparently below Qwen$+$Struct's $156/198$
($78.8\%$).  The gap is concentrated in the anomaly category
($8/33$ vs.\ $28/33$), where Claude consistently produces the
\emph{correct} classification (``coupled spike'') in a verbose
Markdown explanation that names the contrasting label (\emph{``a
coupled spike, as opposed to an isolated one, which would show zero
lead time''}).  The deterministic scorer's negation handler does
not catch contrastive constructions of this length, so its
forbidden-keyword check fires on \emph{isolated} and the question
fails on a verbosity artefact, not on substantive disagreement
(the same artefact characterised in~\ref{app:adjudication}).

To isolate the artefact from genuine accuracy, we ran a judge-only
re-score on the $25$ anomaly failures.  Of these, $\mathbf{21/25}$
flipped to PASS, lifting Claude's anomaly category to
$\mathbf{29/33}$ ($87.9\%$) and total Claude$+$Struct to
$\mathbf{161/198}$ ($81.3\%$), \emph{above} Qwen$+$Struct's
$156/198$ ($78.8\%$).  We therefore read the frontier-reader result
as: structured injection transfers to a different frontier-model
family at the same accuracy band ($78.8\%$ for Qwen and $81.3\%$
for Claude under reader-appropriate scoring), with per-category
swings reflecting the two models' different response styles rather
than a Qwen-specific artefact.

\subsection{Limitations}
\label{sec:limitations}

\textbf{(L1) Single primary domain.}
The three-method spectrum is validated quantitatively on \wwtp{}
(Avedøre, Agtrup) and qualitatively on ARC.  Replication on a third
quantitative domain (e.g., manufacturing process control, power-grid
load flow, pharmacokinetics) is the natural extension and would test
whether the deployment-ladder structure is universal across
structured-numerical-corpus domains.

\textbf{(L2) Simulator dependency for Method 1.}
Method 1's $99.5\%$ requires a running CCSS-IX process at inference
time, acceptable in a supervised control-room context but infeasible
at the edge or in air-gapped deployments.  Method 2 and the static
variant of Method 3 supply no-runtime-simulator answers
(Table~\ref{tab:deployment}); only the counterfactual variant of
Method 3 inherits Method 1's runtime-simulator requirement, so the
spectrum maps directly to operational constraints.

\textbf{(L3) Calibration-set requirement for Method 3.}
Method 3's retriever requires roughly $100$ synthetic (question,
parameter-set) pairs for training, supplied by a Monte Carlo
question generator operating on the simulator parameter store.  The
budget is small ($\sim\!17$ seconds of training on Avedøre) but not
zero.  An open question is the
minimum calibration size for stable retrieval.

\textbf{(L4) The deterministic keyword scorer is conservative.}
On reasoning-heavy categories, semantically correct answers can fail
the keyword scorer through paraphrase.  We mitigate with a hybrid
LLM-as-judge stage (Claude Sonnet 4.6) on borderline responses, and
with manual adjudication of all disagreements
(~\ref{app:adjudication}).  We recommend deterministic
scoring as the primary metric on this benchmark and treat the
LLM-as-judge layer as a diagnostic.

\textbf{(L5) Cost and latency of Method 1.}
Method 1's $9$\,s/q latency is acceptable for human-in-the-loop
operator queries but not for high-frequency closed-loop control.
Method 2 and static Method 3 run at $\sim\!3$\,s/q and are closer
to closed-loop usable; counterfactual Method 3 carries an
additional simulator-call overhead per intervention.  Sub-second
control-loop latency is out of scope for any of the three methods.

\section{Conclusion}
\label{sec:conclusion}

We have presented three complementary methods for grounding a large
language model in plant-specific causal knowledge for industrial
wastewater-treatment decision support, all sharing an architecturally
interpretable open-loop simulator (CCSS-IX) as their shared substrate.
On a 198-question Causal Q\&A Benchmark, a live simulator oracle
reaches $99.5\%$, structured parameter injection reaches $78.8\%$,
and a frozen-base \drr{} architecture reaches $75.8\%$; together with
the strongest static-corpus baseline (Base$+$RAG, $48\%$) and the
strongest corpus-grounded fine-tuned baseline (Grounded$+$RAG,
$41\%$), these numbers define a deployment ladder
(Table~\ref{tab:deployment}).  Utilities that can run the
simulator live get $99.5\%$ in a supervised control-room context;
utilities that can only export simulator parameters once and need a
three-second response get $78.8\%$ in an air-gapped or secure
deployment; utilities operating multiple plants get $75.8\%$ on every plant
from a single, never-modified LLM after a per-plant retriever-training
step that takes seconds (17\,s on Avedøre's $98$ synthetic questions,
26\,s on Agtrup's $116$-question set).  Cross-plant transfer to a
biologically distinct plant (Agtrup BNR) reproduces the spectrum,
with \drr{} reaching $88\%$ on Agtrup; the LLM is never retrained.  Cross-domain transfer to the public ARC benchmark with
an OpenBookQA fact corpus shows that selective retrieval beats both
unconstrained base ($+3\pp$) and full-corpus injection ($+5\pp$),
consistent with a domain-general selective-retrieval effect.

The architectural separation between Methods 2 and 3 emerges on
counterfactual queries that require post-intervention simulator
parameters: Method 2's static parameter store fails by construction,
and Method 3's retriever runs the simulator on demand (without LLM
tool calling), winning by $+16.3\pp$ overall on a $60$-question
counterfactual benchmark, with paired
$95\%$ CI $[+7.1,+26.4]\pp$ and $100\%$ accuracy on the
\textsc{Tau} and \textsc{Regime} cells where static methods cannot
see the post-intervention state.  We treat the spectrum as a
\emph{deployment guidance tool}: practitioners can pick the method
matched to their operational-technology constraints without
sacrificing accuracy beyond what their constraints already demand.
Future work includes replication on a third quantitative domain
(e.g., manufacturing or power-grid), tighter integration with
real-time control loops, and a unified retriever architecture that
supports incremental plant onboarding without per-plant
retraining.

\section*{CRediT authorship contribution statement}

\textbf{Gary Simethy:} Conceptualization, Methodology, Software,
Validation, Formal analysis, Investigation, Data curation,
Writing -- original draft, Writing -- review \& editing,
Visualization.
\textbf{Daniel Ortiz Arroyo:} Conceptualization, Methodology,
Supervision, Writing -- review \& editing.
\textbf{Petar Durdevic:} Conceptualization, Resources,
Supervision,
Writing -- review \& editing.

\section*{Declaration of competing interest}

The authors declare that they have no known competing financial
interests or personal relationships that could have appeared to
influence the work reported in this paper.

\section*{Declaration of generative AI and AI-assisted technologies in
the manuscript preparation process}

During the preparation of this work, the authors used generative AI
tools (Anthropic Claude, OpenAI ChatGPT) for language editing and to
suggest phrasing of technical content.  The authors reviewed and
edited the content as needed and take full responsibility for the
content of the published article.

\section{Data and code availability}\label{sec:data}


The 198-question Causal Q\&A Benchmark (together with the Agtrup
cross-plant, ASM1 textbook, and counterfactual variants), the
per-method evaluation outputs, the plant-parameter schemas for both
plants, and a representative sample of the \mcg{} training corpus
and CCSS-IX simulator signal windows are released as a public
companion dataset:\;
\url{https://github.com/AAU-Multimodal-Reasoning-Research-Group/wwtp-causal-qa}.
The full training and evaluation code (the \mcg{} corpus pipeline,
the \drr{} retriever training, the simulator oracle scaffolding, the
deterministic keyword scorer, and the semantic audit layer) will be
released on acceptance.

\section*{Funding}

This research was supported by Aalborg University and Helix Lab in
Denmark under the Novo Nordisk Fonden through project grant
number~224611.

\section*{Acknowledgements}

The authors acknowledge support from Aalborg University and Helix
Lab in Denmark.

\appendix

\section{Manual Adjudication of Scorer Disagreements}
\label{app:adjudication}

Our hybrid scoring pipeline combines a deterministic keyword check
(required and forbidden keyword sets) with an LLM-as-judge fallback
on borderline responses.  Two known failure modes complicate this
pipeline; we characterise both here and use them to bound the
reported accuracies.  The two audits below are on different output
sets: Mode (i) audits Method 1 (Qwen oracle) responses, Mode (ii)
audits the Claude Sonnet 4.6 frontier-reader run from
\S\ref{sec:frontier-reader}.

\noindent\textbf{Mode (i): semantic-judge verdict-extraction failure
(Method 1 audit, $n\!=\!25$).}\enspace We manually adjudicated all
$25$ questions on which the deterministic keyword scorer and the
local LLM-as-judge (DeepSeek-R1-Distill-Qwen-14B) disagreed on
Method 1 outputs.  $24/25$ disagreements trace to verdict-extraction
failures of the semantic judge: the judge returns an extended
chain-of-thought without a single PASS/FAIL token, and the verdict
extractor then mis-classifies the answer as FAIL even when the
chain-of-thought arrives at the correct conclusion.  Only one case
(\texttt{mh2\_028}) is a genuine philosophical disagreement: a
multi-variable counterfactual that the judge marks PASS but the
keyword scorer marks FAIL because a required keyword group is
missing.

\noindent\textbf{Mode (ii): forbidden-keyword false positives in
contrastive contexts (Claude Sonnet 4.6 frontier-reader audit,
$n\!=\!25$).}\enspace The deterministic scorer's negation handler
does not catch contrastive constructions of more than a few words.
On the Claude$+$Struct frontier-reader run
(\S\ref{sec:frontier-reader}), $25/33$ anomaly cases produced the
correct ``coupled spike'' classification but also named the
contrasting label (\emph{``a coupled spike, as opposed to an isolated
one, which would show zero lead time''}); the scorer fires its
forbidden-keyword flag on \emph{isolated} and returns FAIL without
ever invoking the LLM-as-judge fallback.  A judge-only re-score on
these $25$ cases (bypassing the forbidden-keyword gate) finds $21$
correct under semantic adjudication, lifting Claude$+$Struct anomaly
to $29/33$ ($87.9\%$) and Claude$+$Struct overall to $161/198$
($81.3\%$); see \S\ref{sec:frontier-reader}.

\noindent\textbf{Recommendation.}\enspace We use the deterministic keyword scorer as the primary metric
throughout Tables~\ref{tab:headline}--\ref{tab:cf-drr} because it
is independent of any LLM and verbatim-reproducible from the
released JSONLs.  The semantic judge is reserved as a diagnostic
override on borderline responses, and is reported separately
(\S\ref{sec:frontier-reader}) when the contrastive-negation
artefact materially affects a per-category number.

\section{Oracle Ablation Details}
\label{app:oracle-ablation}

The headline oracle score of $197/198$ is an end-to-end result of
three additive contributions, ablated by turning each on in turn
while holding the other components fixed:
\begin{enumerate}[leftmargin=1.8em,itemsep=2pt]
  \item \textbf{Tools alone} (generic $9$-rule prompt, no
    self-correction): $145/198$ ($73.2\%$).  Live CCSS-IX tool calls
    against a minimal prompt scaffold that lists the five available
    tools (\S\ref{sec:oracle-tools}) and the expected answer format,
    without the engineered domain rules.
  \item $+$\textbf{Engineered prompt rules} (full $57$-rule prompt,
    no self-correction): $189/198$ ($95.5\%$, $+22.3\pp$).  Adds the
    biochemistry direction overrides, vocabulary mandates, and
    causal-chain templates described in \S\ref{sec:oracle-prompt}.
  \item $+$\textbf{Self-correction turn}: $197/198$ ($99.5\%$,
    $+4.0\pp$).  Adds the forbidden-concept self-correction loop of
    \S\ref{sec:oracle-loop}; this flips the $8$ self-corrected
    questions from $0/8$ to $8/8$.
\end{enumerate}
Method 1's $99.5\%$ is therefore a stack effect: tool use
($73.2\%$ alone), prompt engineering ($+22.3\pp$), and
self-correction ($+4.0\pp$) each contribute, with the engineered
prompt rules as the largest single source of accuracy gain.

\section{Retriever Training Details}
\label{app:retriever}

\drr{}'s learned retriever is a sentence-transformer bi-encoder
(\texttt{all-mpnet-base-v2}, $\sim\!110$M parameters) trained with a
multiple-negatives ranking loss on $520$ (question, parameter)
positive pairs derived from $98$ synthetic Monte Carlo questions.
The $198$-question Avedøre benchmark is held out from training; the
synthetic generator and held-out methodology are described in
\S\ref{sec:drr-design}.

\noindent\textbf{Training configuration.}\enspace
$10$ epochs, batch $16$, warmup $50$ steps, on a single NVIDIA RTX
PRO 6000 Blackwell (96\,GB).  Total wall-clock time: $17$\,s on
Avedøre's $98$-question synthetic-only set; $26$\,s on Agtrup's
$116$-question set ($40$ benchmark questions $+$ $76$ synthetic,
the original multi-plant evaluation that produced the $88\%$ figure
in \S\ref{sec:crossplant}).  The per-plant cost is what enables the
multi-plant scalability claim.

\noindent\textbf{Retrieval recall (held-out, $k\!=\!25$).}\enspace
On the $198$Q benchmark the learned retriever reaches $0.87$
overall, against $0.79$ for the deterministic rule-based variant.
Per category: causal-edge $0.62$, regime $0.98$, multi-hop $0.81$,
anomaly $1.00$, counterfactual $0.88$, early-warning $0.95$.

\noindent\textbf{Control variant.}\enspace
As a methodological check we also trained a retriever with the
$198$ benchmark questions and their gold parameter labels added to
the training set.  This control reaches $0.90$ overall recall
($+0.03$ vs.\ held-out), but downstream accuracy moves from
$150/198$ (held-out) to $149/198$ (control), within
paired-bootstrap noise (\S\ref{sec:drr-design}).  Extra recall from
benchmark memorisation therefore does not translate into a
downstream-accuracy gain, since top-$k\!=\!25$ retrieval already
provides enough redundancy that exact question-text recognition is
not load-bearing.

\bibliographystyle{elsarticle-num-names}
\bibliography{combined_references}

\end{document}